# Soft Active EMG Interface for Machine Learning-Enabled Silent Speech Recognition

Yuta Kurotaki[1,2], Shusuke Yamakoshi[1], Reitaro Yoshida[2], Yutaka Isoda[1], Tamami Takano[1], Yuji Isano[1], Yusuke Miyake[2], Kentaro Kuribayashi[2], Hiroki Ota[1*]

[1]Department of Mechanical Engineering, Yokohama National University, 79-5 Tokiwadai, Hodogaya-Ku, Yokohama, Kanagawa 240-8501, Japan
[2]Pepabo Research and Development Institute, GMO Pepabo, Inc., 26-1 Sakuragaoka, Shibuya Ward, Tokyo 150-8512, Japan
*Corresponding author: Hiroki Ota, ota-hiroki-xm@ynu.ac.jp

Co-author Email Addresses: Yuta Kurotaki: kurotaki-yuta-mh@ynu.jp Shusuke Yamakoshi: yamakoshi-shusuke-pn@ynu.jp Reitaro Yoshida: reitaro.yoshida@townwifi.jp Yutaka Isoda: isoda-yutaka-tr@ynu.ac.jp Tamami Takano: takano-tamami-tz@ynu.ac.jp Yuji Isano: isano-yuji-mh@ynu.jp Yusuke Miyake: miyakey@pepabo.com Kentaro Kuribayashi: antipop@pepabo.com

Abstract
Silent speech recognition (SSR) provides an alternative communication pathway in the absence of audible speech. However, conventional approaches are limited by the need for constant facial attachment, privacy concerns, and unstable signal acquisition. Here, we propose a soft, active electromyography (EMG) interface that enables word-level SSR using machine learning. Worn on the hand, the device uses a fingertip electrode that can be positioned near the lips to acquire EMG signals only when needed. The interface integrates liquid metal (LM) interconnects, transparent flexible printed circuit (FPC) electrodes, and elastomer encapsulation to ensure high mechanical stability during finger motion. A deep neural network trained on these stable signals achieved a mean accuracy of 97.2 ± 1.3% across three subjects in classifying a 30-word vocabulary, demonstrating robust linguistic discrimination. Furthermore, real-time drone control validates the practicality of this approach in noisy and privacy-sensitive environments where conventional voice recognition fails. This study highlights the potential of soft, wearable EMG systems as secure and intuitive human–machine interfaces.

## Introduction

Silent speech recognition (SSR) has attracted attention as a human–machine interface (HMI) technology that conveys human intent to machines without audible speech[1]. It offers a communication pathway for use in noisy environments, in situations requiring confidentiality, and for individuals with speech impairments. In the future, it is expected to serve as a foundation for intuitive and secure interactions that do not rely on audible speech.

To realize SSR, conventional approaches have explored optical lip-reading[2,3,4], ultrasonic detection of tongue and lip motion[5,6,7], electroencephalography (EEG)-based language estimation[8,9,10], and infrared sensors[11]. However, optical and ultrasonic methods are constrained by environmental conditions, privacy concerns, and device size. Brain–machine interfaces (BMIs) often capture substantial non-speech information, reducing controllability, while accelerometer and gyroscope sensors[12] impose a significant burden due to their large size. Overall, conventional methods have not achieved the combination of high-precision signal acquisition and wearability or social acceptability required for daily use.

Despite the development of multiple SSR approaches, there remains a pressing demand for devices that enable the daily use of silent speech. To achieve this, it is essential to improve wearability around the face and head while providing a system that minimizes user burden and ensures sufficient recognition accuracy.

As a potential solution to this problem, flexible devices employing soft materials have attracted attention. Reported examples include surface EMG (sEMG) sensors based on PEDOT:PSS/polyurethane composites[13], strain sensors comprising metal thin films and polymer films[14], and triboelectric sensors made of PVC/nylon/sponge[15], all of which provide low invasiveness and skin compatibility. Accordingly, SSR devices based on soft materials are promising for daily use.

However, previous soft-material-based methods require sensors to be permanently attached to the face or throat, which compromises user comfort and affects aesthetics and social acceptance. In particular, a hand-worn architecture that contacts the face only briefly and intentionally may offer advantages over face- or throat-mounted devices for users who prefer not to wear conspicuous sensors on the face for extended periods[16,17,18]. More critically, such continuously worn devices may acquire biosignals without the user's intent, leading to risks of privacy invasion and security concerns[19]. Thus, while continuous sensing has often been considered an advantage for wearables, it becomes a limitation in the context of SSR.

To address challenges in aesthetics, privacy, and security, SSR requires not only flexibility and wearability but also a novel operating modality that functions on demand under user control. Rather than serving as a universal replacement for continuous recognition systems, this on-demand paradigm is particularly suited to command-based HMIs, privacy-sensitive interactions, and operation in noisy environments, where user-controlled sensing initiation offers distinct advantages. This concept is relevant not only to SSR but also to smartphones that exploit location histories and healthcare devices that process biometric information, both of which require strict consideration in handling sensitive personal data[20,21].

In this study, we developed a wearable soft, active electromyography (EMG) interface that operates based on intentional user input. The device is worn on the hand, and EMG signals are acquired only when needed by placing a fingertip electrode near the lips. Structurally, the system integrates a flexible elastomer (Ecoflex), a transparent FPC substrate, and highly conductive yet stretchable LM interconnects, enabling stable electrical connectivity under repeated, complex finger movements. This high mechanical stability ensures consistent and reproducible EMG signals, supporting accurate speech recognition through machine learning algorithms.

Moreover, by eliminating the need for continuous facial attachment, the device gives users control over when signals are acquired, thereby enhancing privacy and security. We further demonstrated its utility through SSR and drone operation in noisy environments. The developed soft EMG interface enables both machine-learning-compatible, stable signal acquisition and user-driven, on-demand operation. Evaluation across three subjects showed a mean classification accuracy of 97.2 ± 1.3% for a 30-word vocabulary, establishing a new paradigm for practical SSR.

## Results

### Design for Intent-Driven SSR

The developed wearable silent speech interface, mounted on the hand, enables language recognition and machine control through mouth movements alone, without audible speech. By placing the fingertip near the mouth only when intended (Fig. 1a), the exposed electrode on the fingertip pad contacts the area around the lips, enabling acquisition of EMG signals from the perioral muscles. These signals are then processed by machine learning algorithms to enable character recognition and control of smart devices such as drones (Fig. 1b).

Compared to devices that require sensors to be continuously attached to the face, this device operates only when the user intentionally touches the mouth area with the fingertip, performing language recognition or machine input only at that moment. As the device is physically separated from the perioral skin except during input, no silent speech is sensed outside intended use, and external attempts to trigger sensing by third-party attacks are prevented. This addresses privacy and security concerns and mitigates aesthetic issues for users in cultural contexts where attaching sensors to the face or throat is undesirable. Additionally, as a feature of silent speech, words can be recognized in situations where one wishes to remain quiet and avoid being overheard. No audible sound is used; therefore, device operation remains unaffected even in noisy environments.

The EMG electrode attached to the fingertip comprises a transparent FPC substrate and Ecoflex, with the electrode arranged on the inner surface of a tubular fingertip structure. An EMG amplification circuit is mounted on the dorsum of the hand and connected to the fingertip electrode via LM interconnects. As the finger is structurally complex and highly stretchable, LM wiring is used because it offers superior electrical conductivity, minimal resistance fluctuation under deformation, and stable resistance under repeated strain compared with conventional silver paste, PEDOT:PSS, and carbon-based conductors. By employing LM wiring between the fingertip and the back of the hand, EMG signals can be stably transmitted to the circuit even during repeated finger bending. This technical feature is crucial when incorporating machine learning, which requires large amounts of training data, into stretchable devices[22] (Fig. 1c).

An example of device usage is shown in Fig. 1d. The device is worn on the hand, the fingertip is placed near four muscles surrounding the lips, and EMG signals are acquired from these perioral muscles. Silent speech is then recognized from these signal patterns using machine learning.

### Device Design and Characterization

An active EMG interface was fabricated that combines high skin conformability with mechanical flexibility. The device body employs Ecoflex, a soft elastomer, as the packaging layer, forming a structure that can be mounted on the dorsum of the hand (Fig. 2a). An EMG circuit for amplification and signal processing (Supplementary Fig. 1) is implemented on the Ecoflex substrate, while surface electrodes for direct skin contact are placed on the inner side of the fingertip (Supplementary Fig. 2).
The EMG electrodes were fabricated as a bipolar pair on a transparent FPC substrate. The reference electrode was connected to the REF terminal at the distal end of the circuit board; when the device is worn, it contacts the skin on the dorsal side of the wrist, where tendons and bone are superficial, minimizing local EMG contamination (Supplementary Fig. 3). The interelectrode distance was set to 5 mm to enable EMG measurement within the limited fingertip area. The electrodes were soldered onto copper foils on the transparent FPC substrate to facilitate skin contact. They were adhered to the Ecoflex substrate using Sil-Poxy and then encapsulated with Ecoflex. The EMG electrodes and circuit were interconnected using LM (Supplementary Fig. 4). Electrode–skin impedance was characterized by potentiostatic electrochemical impedance spectroscopy, yielding a mean impedance of 470 ± 103 kΩ at 10 Hz (Supplementary Fig. 5). The device was designed for repeated use and removability; tensile testing of the fabricated electrodes confirmed that the resistance of the electrode–circuit interconnection via LM remained stable ($\Delta R/R_0 < \pm 0.5\%$) across strain levels of 10% to 50% and after repeated cycling (Supplementary Figs. 6 and 7). Finally, the entire structure was encapsulated by spray coating Ecoflex. Owing to its planar structure, the device can be folded around the fingertip, positioning the EMG electrodes on the finger pad to form the interface (Fig. 2b).
The interface was designed to conform closely to the fingers, with fingertip electrodes functioning as dry electrodes. When placed on the perioral muscles, these electrodes detect weak EMG signals associated with speech articulation with high sensitivity. The acquired EMG signals are amplified within the EMG circuit, passed through low-pass and high-pass filters to remove noise, subjected to analog-to-digital conversion (ADC), and transmitted to an external computing device for real-time processing (Fig. 2c).
The fingertip electrodes were anatomically positioned to correspond to four speech-related muscles: the buccinator, orbicularis oris, depressor labii inferioris, and mentalis. The buccinator stabilizes the cheeks during plosive and fricative production. The orbicularis oris is responsible for lip rounding and protrusion, the depressor labii inferioris moves the lower lip, and the mentalis acts in coordination with movements that retract the mouth corners. By aligning the electrodes with these muscles, EMG activity associated with speech articulation could be effectively recorded (Fig. 3a).
An example EMG waveform of the mentalis muscle is shown for a single channel during speech and rest (Fig. 3b). The signal shows EMG activity during 1 s of silent speech, followed by 1 s of rest and another 1 s of silent speech. In the amplification circuit used in this study, muscle activity is captured within the range 0–3.3 V, centered at 1.6 V. This 1.6 V bias originates from the single-supply analog front-end, where a shunt voltage reference (LM4051) shifts the bipolar EMG signal into the ADC input range (0–3.3 V). Although no digital high-pass filter is applied before feature extraction, the DC component is structurally eliminated within the MFCC pipeline: the FFT confines DC energy to the 0 Hz bin, the Mel filter bank covers frequencies above approximately 20 Hz and therefore excludes the 0 Hz bin, and the subsequent DCT yields MFCC coefficients ($c_0$–$c_6$) that contain no DC-derived content. Consequently, the 1.6 V offset does not affect the classification features.
Using this device, we conducted experiments to measure muscle activity during silent speech. Figure 3c presents four-channel EMG signals recorded during silent articulation of the words *"Move"* and *"Turn,"* selected from a 30-word vocabulary. The displayed waveforms were high-pass filtered to remove low-frequency baseline drift, thereby revealing the physiological EMG bursts associated with articulation more clearly. The signals correspond to the buccinator, orbicularis oris, depressor labii inferioris, and mentalis muscles, reflecting localized muscle activity associated with articulation. Distinct EMG patterns were observed between *Move* and *Turn*, with differences corresponding to coordinated movements of the respective muscles. These results suggest that the proposed device can capture input signals sufficient to distinguish between intended words, thereby functioning as an effective silent speech interface.

## Implementation of Machine Learning-Integrated Silent Speech Interface

Using the high-conformability EMG device described in the previous section, we constructed a machine learning model for word recognition during silent speech and achieved classification of voiceless articulation. During silent speech, fingertip electrodes were placed near four perioral muscles, and EMG signals were acquired at a sampling rate of 1 kHz (1 ms intervals). From the four-channel EMG signals, Mel-frequency cepstral coefficients (MFCCs) were computed and used as inputs to a deep neural network (DNN) trained on EMG-derived MFCC features (Fig. 4a). The architecture of the DNN (Supplementary Fig. 8) comprised multiple fully connected (dense) layers following the input layer, with dropout and batch normalization applied to each layer. The number of hidden layers, units per layer, dropout rate, and learning rate were determined through systematic hyperparameter optimization using Keras Tuner with the Hyperband algorithm over a broad search space. This design stabilized training and suppressed overfitting.

The acquired EMG signals were preprocessed by framing into fixed time windows, followed by application of a fast Fourier transform (FFT) to each frame. The resulting frequency components were processed through a Mel filter bank and subjected to a discrete cosine transform (DCT) to obtain MFCCs. The first- through sixth-order MFCCs were extracted as features and used as inputs to the DNN (Fig. 4b).
EMG signals were acquired from three subjects (S01, S02, and S03). Subject S01 contributed 60 trials per word (N = 1,800 total), while S02 and S03 each contributed 30 trials per word (N = 900 total). The training sets for S02 and S03 were augmented to 60 trials per word via Gaussian jitter to match S01. Using a combination of input channel selection, stratified data splitting, cross-validation, callbacks, hyperparameter tuning, and normalization, a mean test accuracy of 97.2 ± 1.3% was achieved across the three subjects (S01: 97.5%, S02: 98.3%, S03: 95.8%) (Fig. 4c). Detailed classification metrics are provided in Supplementary Table 3 and Supplementary Fig. 9, and per-class confusion matrices for all subjects are presented in Supplementary Fig. 10. Supplementary Fig. 9 shows that train–test gaps remained within 1.7–4.2 percentage points across all subject-dependent models and that test accuracy closely matched the macro-averaged F1 score for each subject, confirming balanced classification across all 30 word classes. The combined cross-subject model achieved a test accuracy of 92.1% and a macro F1 score of 0.921, demonstrating robust generalization despite inter-subject variability in EMG patterns.
Supplementary Table 3 summarizes the five-fold cross-validation scores, held-out test accuracy, and macro-averaged F1 scores for all three subject-dependent models and the combined cross-subject model. The per-class confusion matrices in Supplementary Fig. 10 exhibit strong diagonal dominance across all 30 words for each subject, with nearly all classes achieving 12/12 correct classifications. This indicates that the model effectively discriminates the target vocabulary using perioral EMG signals acquired via fingertip dry electrodes, without systematic confusion between specific word pairs.
We further examined whether adjusting the number of input channels improved feature learning. Performance with three input channels is shown in Supplementary Fig. 11; the four-channel configuration achieved the best results overall. Generalization performance was evaluated using a stratified train–test split (test_size = 20%, stratified by class label), ensuring that exactly 12 samples per class were assigned to the test set. This stratified partitioning yields a uniform class distribution across all 30 word classes in both the training set (N = 1,500) and the test set (N = 300), providing an unbiased estimate of model performance. In addition, five-fold cross-validation was applied to enhance the reliability of model evaluation. By repeatedly using the entire dataset for both training and testing, variance due to specific splits was reduced, enabling a more robust estimation of performance.
For classification results, t-distributed stochastic neighbor embedding (t-SNE) was applied to visualize the feature space, confirming that separability between words improved after application of the machine learning model (Figs. 4d and e). These results demonstrate that the proposed approach is effective for EMG-based silent speech interfaces and highlight its practicality as a non-audible communication method.

## Application to Human–Computer Interaction: Demonstration via Drone Control

To evaluate the applicability of the developed high-conformability EMG device, we conducted a proof-of-concept experiment in which silent speech was used for real-time word classification and drone control. Participants wore the device and articulated commands without producing audible speech. The resulting EMG signals were analyzed and classified in real time, enabling recognition of control inputs (Fig. 5a). The device was connected to a microcontroller, while the trained model was executed on a PC for real-time classification.
For each of the four EMG channels, six-dimensional MFCCs were computed as features (Fig. 5b). The signals were segmented into short time windows, transformed to the frequency domain using the short-time Fourier transform (STFT), and mapped through a Mel filter bank. A DCT was then applied to the logarithmic Mel spectrum, and the six lowest-order MFCCs from each channel were used as inputs.
Using these features, we implemented a control scheme based on four commands: *Start*, *Move*, *Turn*, and *Stop*. A small quadcopter (Tello) was used to demonstrate feasibility, with commands transmitted over Wi-Fi (Fig. 5c). Recognition of *Start* initiated autonomous takeoff and hovering. Once airborne, *Move* prompted forward motion over a preset distance. Recognition of *Turn* caused the drone to rotate by 45° either clockwise or counterclockwise, emulating joystick-based yaw control. Finally, recognition of *Stop* after the flight sequence instructed the drone to land automatically. All commands were classified in real time from EMG-derived features and transmitted wirelessly (Fig. 5d).
As a result, takeoff, translation, turning, and landing were executed using only silent speech inputs, without audible sound (Fig. 5e). These findings demonstrate that the proposed device supports real-time operation with high classification accuracy and functions effectively as a silent speech interface. The results further highlight its utility as an intuitive, nonverbal method for human–machine interaction in environments where speech is impractical or confidentiality is required. A demonstration is provided in Supplementary Video 1.

## Discussion

In EMG measurements for SSR, particularly for daily use, electrode portability and reusability remain critical challenges. To address this, we developed EMG measurement electrodes and amplification circuits that prioritize portability and reusability. Both wet and dry electrodes are available for acquiring[23] EMG signals from the perioral skin, and the choice depends on the application. Gel-based electrodes[24] and microneedle electrodes[25,26] exhibit high adhesion. Although gel electrodes provide secure fixation, they require gel application to the skin, which is unsuitable for repeated attachment and removal. Microneedle electrodes are invasive and therefore unsuitable for daily use involving repeated skin contact. Accordingly, we adopted an active EMG approach to enable noninvasive signal acquisition from the skin surface.

Active EMG allows noninvasive measurement without gels. However, such electrodes typically require fixation using adhesive tape or bands; insufficient fixation increases susceptibility to noise and hinders stable EMG acquisition[27]. Electrodes based on PEDOT:PSS[13,28], carbon nanotubes or carbon black[29,30,31], and silver paste[32] have been reported. Although reusable PEDOT:PSS electrodes have been investigated[33], their resistance varies with strain and temperature. This makes them unsuitable for the present study, in which the electrodes must withstand pressure during manual placement near the face and remain stable under environmental variations when worn on the hand.

For carbon- and silver-based systems, prior work has compared cPDMS electrodes (carbon–PDMS), AgPDMS electrodes (silver–PDMS), and stainless-steel electrodes[34], and the material characteristics of dry electrodes have also been evaluated[35]. Based on these findings, we adopted solder-coated electrodes that connect readily to copper foil on transparent FPC substrates and exhibit minimal resistance variation under strain.

Ecoflex was applied around the electrode contact area to prevent slippage during fingertip-based manual fixation against the perioral muscles. By combining Ecoflex as a flexible base material with transparent FPC electrodes and LM interconnects that accommodate finger motion, we designed a system that allows users to intentionally press the electrodes against the skin for stable placement on target muscles (Fig. 2b). This structure eliminates the need for adhesives or tape while maintaining stable fixation, thereby overcoming the limitations of conventional dry electrodes. Signal quality was further validated under three levels of applied contact pressure, confirming stable EMG acquisition (SNR > 30 dB) across all conditions (Supplementary Fig. 12). Thus, the system combines ease of use with reusability. The EMG amplification circuit incorporates both low-pass and high-pass filters, enabling extraction and amplification of weak EMG signals while removing environmental noise and unwanted frequency components (Fig. 2c).

Although the fingertip electrodes use solder-coated metallic contacts with conventional dry-electrode characteristics, the overall interface integrates soft elastomer encapsulation and LM interconnects. As a result, while the electrode surface itself is not intrinsically soft, the device as a whole provides high mechanical compliance and wearability, justifying its designation as a “soft active EMG interface.”

For silent speech classification using machine learning, electrodes were placed on the perioral region without skin pretreatment. Participants repeated a cycle of 1 s of rest followed by 1 s of silent speech, during which EMG signals were acquired. In single-channel measurements of the mentalis muscle, although noise was present, voltage increases corresponding to muscle activation during speech movements were observed (Fig. 3b). During articulation of *Move*, characterized by lip protrusion, EMG signals were recorded in channel 3 (depressor labii inferioris) and channel 4 (mentalis). These muscles are closely involved in lip protrusion and closure[36]. The depressor labii inferioris lowers the lower lip, while the mentalis protrudes it forward. Detection of their activity during lip pursing supports the anatomical validity of the measurement. Similarly, during silent articulation of *Turn*, EMG signals were recorded for approximately 0.75 s in channel 2 (orbicularis oris) and channel 4 (mentalis) as the lips closed. The orbicularis oris functions in closing and rounding the lips, while the mentalis raises the lower lip and wrinkles the chin, moving the lower lip upward and forward[37]. These results indicate that relevant muscle activity during silent speech was successfully captured.

These findings demonstrate that distinctive features of words can be detected during silent speech. Articulatory gestures such as bilabial closure (plosives), constriction (labiodental fricatives), and rounding (approximants) are widely observed across languages. According to the UCLA Phonological Segment Inventory Database, 446 of 451 languages (99%) include bilabial plosives, and many also feature labiodental fricatives and rounded approximants[38]. This suggests that the results of the present study may extend beyond English to other languages (Fig. 3c).

By integrating EMG signals acquired using the proposed device into a machine learning model, word classification accuracy for silent speech reached 97.2 ± 1.3% (Fig. 4c). This result indicates that speech can be distinguished based on the activity of four perioral muscles. In contrast to previous studies that directly processed EMG signals, our model applied a feature transformation into MFCCs, a representation widely used in speech recognition, and used these as training data (Figs. 4a and b; Supplementary Fig. 13). This approach has proven effective in EMG-based SSR research[39,40,41] and similarly improved classification accuracy in this study. The use of MFCCs with convolutional neural networks (CNNs) has been reported to achieve higher accuracy than classifiers such as support vector machines (SVMs)[41], and this perspective informed our model selection.
We also evaluated a Conformer model[42] based on Transformer architectures[42,43], which achieved an accuracy of 91.1%. In comparison, the MFCC-based DNN achieved the highest accuracy of 97.2% (Supplementary Fig. 14). This result suggests that under the limited data conditions of this study, the simpler DNN provided better generalization than the more complex Conformer model. Training dynamics across all subjects showed that both training and validation losses converged within approximately 20 epochs, after which the validation loss plateaued at a low level without increasing through 150 epochs, indicating the absence of overfitting (Supplementary Fig. 15).
The device is designed to be worn on the hand, with electrodes placed on the fingertips to measure perioral muscle activity. Owing to anatomical constraints, the maximum number of electrodes is five. In this study, we evaluated four electrodes placed from the thumb to the ring finger, achieving a classification accuracy of 97.2 ± 1.3%. This result demonstrates that high-performance recognition can be achieved without adhesives and with a limited number of channels. Previous studies have shown that increasing the number of electrodes, for example to 7, 8, or 12, provides more feature information and can improve classification accuracy[39].
In our system, a practical constraint is that users manually press the device against the mouth region. This creates a dynamic measurement environment rather than a fully static one, with noise and signal variability arising from electrode placement and hand movement. Because the training data were collected under these natural, non-standardized contact conditions, the DNN implicitly learned to accommodate variability associated with fingertip-based electrode placement. Despite these constraints, including the use of only four electrodes and dynamic operation, the device achieved a high classification accuracy of 97.2 ± 1.3%. Comparisons using two- and three-electrode configurations also showed that four electrodes yielded the highest accuracy, consistent with prior findings that accuracy improves with additional electrodes (Supplementary Table 1).
Furthermore, the classification accuracy achieved in this study is high compared with previous studies that assumed fixed electrodes under static measurement conditions[39]. Visualization using t-SNE confirmed that silently articulated words can be distinguished in the feature space. This result demonstrates the feasibility of SSR based on EMG signals, consistent with prior studies that used t-SNE to evaluate classification of movements[44]. These findings support the effectiveness of the proposed device and preprocessing approach (Figs. 4d and e).
As a practical application of the proposed wearable silent speech interface, we conducted drone operation experiments. Conventional voice recognition systems are susceptible to interference from loud operational sounds, such as drone startup and flight noise, as well as background noise in environments such as industrial or disaster sites, making it difficult to recognize soft or whispered speech[45,46]. In contrast, the present system performs SSR based on EMG signals and does not require microphone input. As a result, command-based control remained feasible during the experiments, unaffected by environmental noise or drone operation. This capability allows the system to overcome limitations of conventional speech recognition in noisy environments and function as a robust human–machine interface. Specifically, EMG signals acquired by the device were processed by a pretrained silent speech classification model, which recognized commands and enabled drone control.
The commands used were *Start*, *Move*, *Turn*, and *Stop*, which are commonly employed in smart device operation. Silent speech-based drone control has been explored in prior research[47]; in this study, we extended its applicability by mapping classification outputs to control commands and transmitting them to the drone in real time. Experimental results demonstrated that a complete sequence of operations, including takeoff, movement, turning, and landing, could be executed continuously.
Supplementary Table 2 summarizes smart devices capable of SSR via surface sensing of the face, including EMG-based lip-reading[48], transparent facial patch sEMG[49], and jaw-motion-based authentication[50]. A recent review has further surveyed emerging unvoiced-speech wearable approaches and their broader design considerations[51].
These systems typically assume continuous attachment and may raise privacy concerns due to unintended acquisition of muscle activity. In contrast, the proposed design initiates signal acquisition only when the user deliberately brings the hand close to the mouth. This approach transfers control of information acquisition to the user and introduces a paradigm that addresses both privacy and security concerns.
Regarding classification performance across users, subject-dependent models trained and evaluated on individual participants (S01, S02, and S03) achieved a mean test accuracy of 97.2 ± 1.3%, demonstrating consistent performance. A combined cross-subject model trained on all three subjects achieved a test accuracy of 92.1% and a macro F1 score of 0.921 (five-fold CV score: 0.836) (Supplementary Fig. 16). Across all models, held-out test accuracy consistently exceeded the CV score, indicating that the test sets were representative and that the models did not overfit to the training data. The performance gap between individual and combined models is attributable to inter-subject variability in EMG patterns; accordingly, per-user calibration is recommended to maximize performance.

Traditional silent speech interfaces have primarily been studied to protect communication privacy. However, many patch-type EMG systems[52,53], while capable of stable measurement, require electrodes to be attached over large areas of the face or neck. As a result, EMG signals are continuously acquired even when not intended, raising privacy concerns. Systems that detect laryngeal muscle motion using magnetoelasticity[54] have also been investigated, but they still require continuous attachment using tape or similar fixation methods and therefore share the same limitation. In addition, approaches that use cameras worn between the neck and chest to recognize mouth movements[2] introduce privacy risks because unintended content may be captured.

In EEG- and neural network-based studies[10], participants are required to wear caps, resulting in continuous acquisition of brain activity. SSR has also been explored using everyday objects such as glasses, masks, or earphones, integrated with accelerometers[55], infrared sensors[11], TENGs[15], and strain sensors[44]. Although these devices offer high wearability and a familiar appearance, they continuously capture mouth movement, raising concerns about privacy. In addition, personal information and behavioral patterns may be collected without user consent; thus, always-on wearable sensors inherently carry risks of unintended speech capture and potential security threats from external attacks. These issues arise because the devices are physically affixed to the body. Accordingly, alternative approaches that do not require continuous facial attachment have been investigated for daily use[56].

Ultrasound echo-based silent speech devices[6] enable intentional placement against the throat during measurement; however, they require gels and are not well suited for daily use, and their handheld operation limits portability. In contrast, the present device is designed as an integrated system that does not require continuous facial attachment. By employing soft materials, it enables EMG acquisition only when deliberately pressed against the mouth, without the need for gels or adhesive tape.

This design restricts data acquisition to intentional use, thereby mitigating privacy concerns. Moreover, briefly touching the perioral region with a fingertip is a natural and common gesture, making the device inconspicuous during use compared with face- or throat-mounted sensors that remain visibly attached. Although the measurement conditions are less controlled than those of adhesive-based electrodes, because no tape fixation or skin pretreatment is used, the device achieved a silent speech classification accuracy of 97.2% when combined with machine learning. This performance is comparable to that of related systems. Signal quality was further supported by SNR measurements under varying contact pressures, yielding values of 31.0–35.5 dB across low, medium, and high pressure conditions (Supplementary Fig. 12). These values are consistent with or exceed the 20–30 dB range typically reported for dry-electrode surface EMG systems, confirming that the fingertip-based electrode configuration provides sufficient signal fidelity for robust classification. Drone control experiments further demonstrated the practicality of the device as an HMI.

In this study, we developed a user-controlled wearable soft EMG interface that provides stable signal acquisition and enables SSR. By integrating LM interconnects with elastomer encapsulation, the device maintained reliable electrical performance under repeated finger movements and delivered consistent biosignals suitable for machine learning. Using a DNN, the system classified a 30-word vocabulary with an accuracy of 97.2%, achieving word-level recognition without continuous facial attachment.

Furthermore, drone control experiments in noisy and privacy-sensitive environments verified its utility for intuitive human–machine interaction. The significance of this study lies in three contributions: (1) a soft electrode–amplifier interface combining Ecoflex encapsulation and LM interconnects for mechanically compliant EMG acquisition, (2) an on-demand sensing paradigm that provides a physical-layer privacy guarantee by restricting signal acquisition to intentional user contact, and (3) integration of these elements into a machine-learning-compatible SSR system that achieved 97.2% accuracy across 30 words without continuous facial attachment. Together, these contributions establish a new paradigm that addresses challenges related to privacy, social acceptability, and signal stability in wearable SSR systems.

This work highlights the potential of soft and stretchable bioelectronics as safe and intuitive communication tools. Future work should include systematic evaluation of the impact of electrode placement variability on recognition accuracy, development of methods to mitigate these effects, expansion of the vocabulary, and extension to continuous sentence-level recognition using sequence modeling approaches such as connectionist temporal classification (CTC) or sequence-to-sequence architectures. Addressing these challenges would enable this technology to evolve into a broadly applicable platform for communication, assistive devices, and control systems in real-world environments.

# Methods

## Fabrication of EMG amplification circuit board

The EMG measurement circuit comprised an instrumentation amplifier (AD627, Analog Devices) at the front end and two operational amplifiers (AD8607, Analog Devices) in the subsequent stage to amplify weak potentials. High-pass and low-pass filters were incorporated to extract relevant EMG signals, following established designs for biological signal processing circuits[57]. The reference voltage was set to 1.6 V, enabling measurement of muscle potential changes within the range 0–3.3 V (Supplementary Fig. 1). The circuit board was fabricated using a transparent polyimide substrate coated on one side with copper foil (L71KTS 1012EDR T10, Arisawa Mfg. Co., Ltd.). AZ photoresist was spin coated, dried on a hot plate at 100 °C, and patterned by mask exposure. The circuit pattern was then developed by etching (Supplementary Fig. 2), after which the required components were mounted. Two measurement electrodes and one reference electrode were applied to the skin for signal acquisition.

## Preparation and wiring of LM paste

LM paste was prepared as follows. Nickel powder (3–7 µm, Alfa Aesar Co.) was dispersed in Galinstan (Maruya.com) at mass ratios of 2% and 5%. The mixture was sonicated using an ultrasonic probe (SFX 550, BRANSON) with a duty ratio of 30% and a total energy input of 6 kJ. It was then exposed to air overnight to promote oxidation. Oxidized LM paste was also prepared by stirring Galinstan under ambient conditions using a stirrer (Azone) at 750 rpm for 60 min to induce oxidation.

## Fabrication of silent speech device

A mold produced via 3D printing was filled with silicone rubber Ecoflex 00-20 (Smooth-On; A:B = 1:1 by weight) and cured in an oven. The EMG amplification circuit board was fixed onto the cured elastomer using Sil-Poxy adhesive (Smooth-On). Fingertip electrodes were attached in the same manner using Sil-Poxy. A polyimide film was cut using a laser marker (Keyence) to form a stencil mask for LM wiring. LM paste was applied through the mask, and after removal of the mask, wiring connections between the fingertip electrodes and the EMG circuit were formed. The connection areas between the circuit board and the LM wiring were sealed with silicone rubber (KE-1606, Shin-Etsu Chemical) and cured. Finally, Ecoflex was coated over the entire device and cured in an oven to encapsulate the system.

## Software for silent speech classification

A microcontroller (Arduino) was used to digitize signals obtained from the electrodes and EMG amplification circuit. EMG signals from four perioral muscles were sampled at 1 ms intervals (1,000 Hz) via four analog input channels and transmitted to a computer (Fig. 2c).

EMG data were collected from three subjects (S01, S02, and S03). Subject S01 performed 60 trials per word (1,800 samples across 30 words), while S02 and S03 each performed 30 trials per word (900 samples total). To equalize trial counts, training sets for S02 and S03 were augmented to 60 trials per word using Gaussian jitter, with a noise scale of 2% of the per-channel standard deviation, applied to raw EMG signals before MFCC extraction. Augmentation was applied only to the training sets, while test sets consisted of unaugmented recordings.

The dataset was partitioned into training and test sets using a stratified split (test_size = 20%, stratified by class label), ensuring exactly 12 samples per class in the test set. Five-fold cross-validation was applied to improve the reliability of model evaluation and to support hyperparameter optimization using Keras Tuner. These data were used to construct test datasets for prediction with machine learning models.

The EMG signals were transformed into MFCCs, which served as input features for machine learning. A five-layer DNN was used for training, with the final layer employing a softmax function for multiclass classification of silent speech. The classification results were further validated using t-SNE, confirming that silently articulated words could be distinguished in the feature space (Figs. 4d and e).

## Drone operation application

A control program was implemented in Python. The PC connected to the device communicated with a drone (Tello, Ryze Technology) via Wi-Fi. During silent speech articulation, with the device positioned near the mouth, recognized words were transmitted via HTTP to control the drone (Fig. 5c). The recognized words were displayed in a web browser, and the drone executed movements within the room based on these commands (Figs. 5d and e).

## Ethical approval and participant consent

All participants provided informed consent after receiving a full explanation of the study's purpose and procedures. The research protocol was approved by the Ethics Committee of the Yokohama National University Graduate School of Engineering Science (No. 2020-16; approved February 12, 2021).

## Author contributions

YK conceived the study and was responsible for the overall design, device implementation, and experimental validation. SY contributed to the device fabrication processes. RY performed the machine learning analyses. YI and TT contributed to the circuit design. YI (Yuji) contributed to experimental planning for device fabrication. YM and KK contributed to machine learning analyses and experimental planning. HO supervised the project and provided overall guidance. All authors read and approved the final manuscript.

## Acknowledgements

This work was supported by JST AIP Acceleration Research (JPMJCR22U2), Japan, JSPS KAKENHI Grant Number JP24H00296, and MEXT KAKENHI Grant Number JP24H00887. The funding bodies provided financial support but had no role in the design of the study, the collection, analysis, or interpretation of data, or in writing the manuscript.

## Competing interests

All authors declare no financial or non-financial competing interests.

## Data availability

The datasets generated and/or analysed during the current study will be deposited in a publicly accessible repository with a persistent URL. Access details will be provided upon publication.

## Code availability

The underlying code and training/validation datasets for this study are planned to be released in a publicly accessible repository with a persistent URL. The link will be provided at the time of publication.

**Figure 1: Intent-driven SSR. a**, Conceptual diagram of the proposed wearable silent speech interface. Electrodes are mounted on the hand, and the hand is brought near the mouth only when needed to acquire EMG signals. **b**, EMG signals are acquired by positioning the hand around the mouth during silent speech, and a DNN recognizes the signals to operate the device. **c**, The device is worn on a finger and integrates LM wiring, Ecoflex, and transparent FPC electrodes to accommodate hand and mouth movements. **d**, The device is used by placing the fingertip near the mouth. It includes four electrodes for EMG acquisition, positioned around the perioral region.

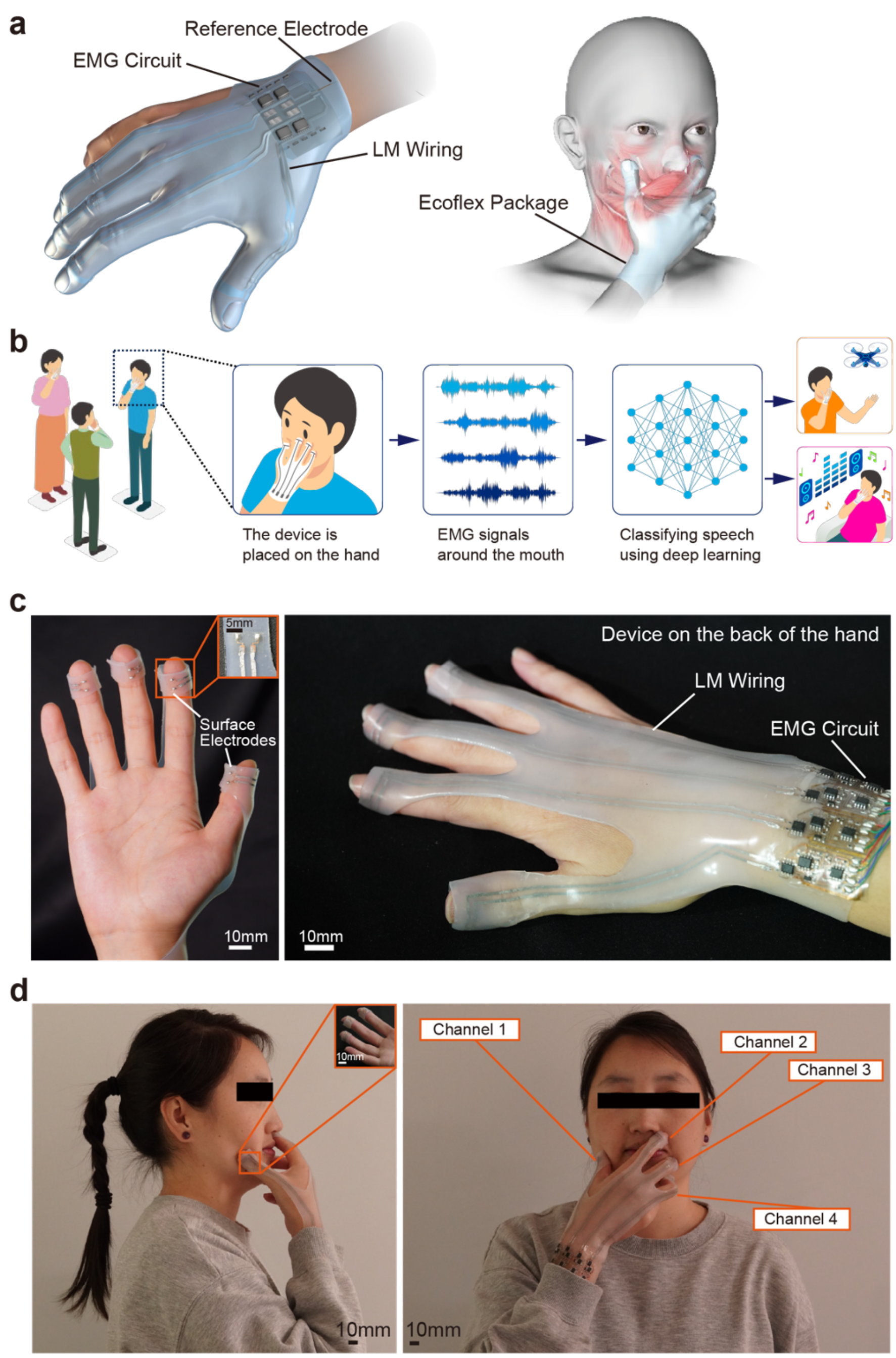
a
Reference Electrode
EMG Circuit
LM Wiring
Ecoflex Package
b
The device is placed on the hand
EMG signals around the mouth
Classifying speech using deep learning
c
5mm
Surface Electrodes
10mm
Device on the back of the hand
LM Wiring
EMG Circuit
10mm
d
10mm
Channel 1
Channel 2
Channel 3
Channel 4
10mm
10mm

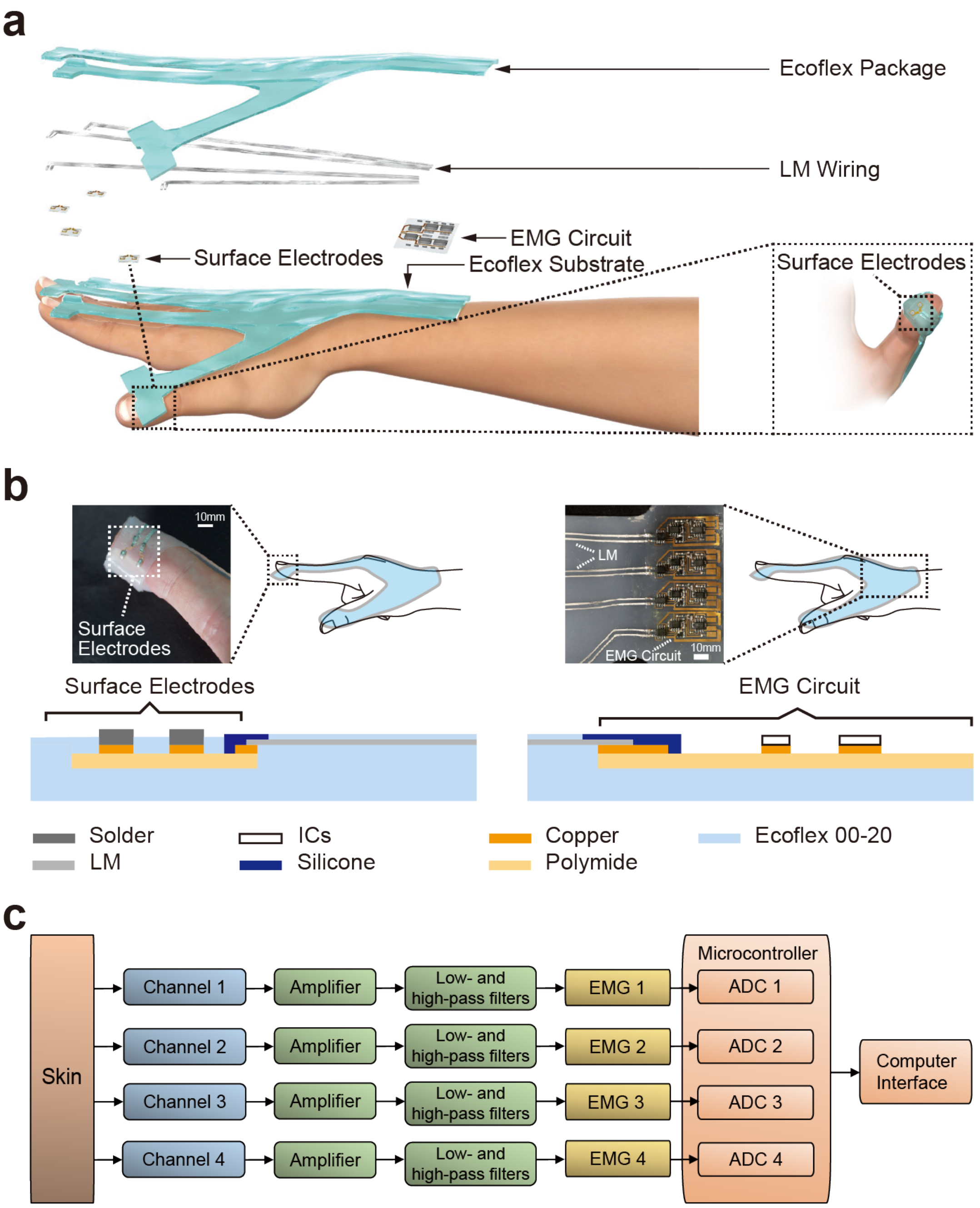

**Figure 2: Device structure of the soft active EMG interface and data processing. a**, Electrodes are mounted on an Ecoflex substrate for attachment to the fingertip. LM wiring connects the electrodes to the EMG circuit, which is encapsulated with Ecoflex. **b**, Cross-sectional view of the device, showing electrodes positioned on the fingertip for skin contact during silent speech and the EMG amplification circuit located on the wrist side. **c**, Flowchart illustrating signal acquisition and processing: four electrodes are placed on the skin, EMG signals are amplified, passed through low- and high-pass filters, and converted to digital signals to capture movements around the lips.

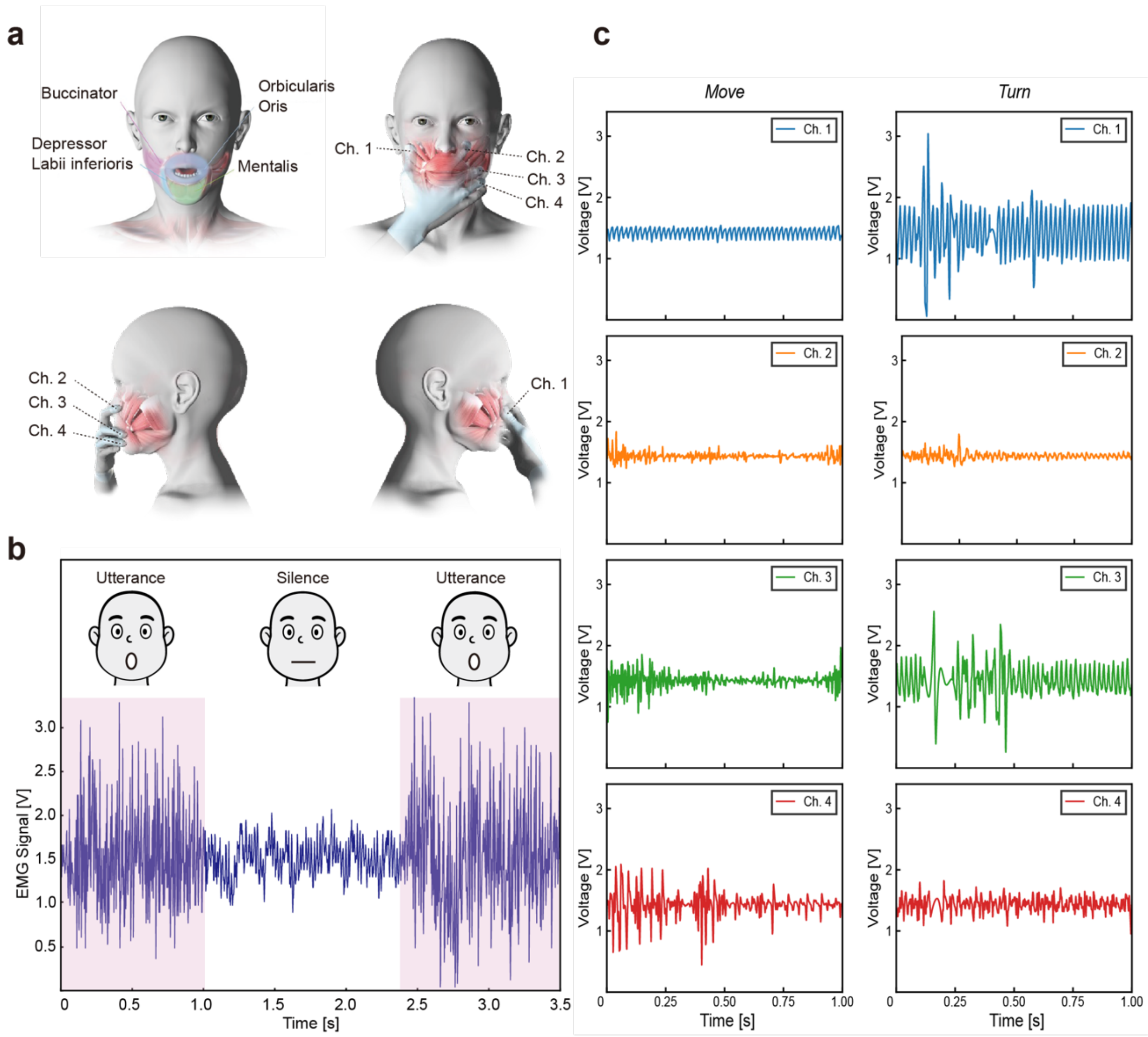

**Figure 3: Electrode interfaces for EMG recording. a**, Placement of fingertip electrodes on four lip-related muscles: the buccinator, orbicularis oris, depressor labii inferioris, and mentalis. **b**, Representative EMG waveform of the mentalis muscle, showing alternating periods of rest and silent articulation. **c**, Four-channel EMG signals recorded during articulation of "*Move*" and "*Turn*," highlighting distinct muscle activation patterns associated with different phonetic gestures.

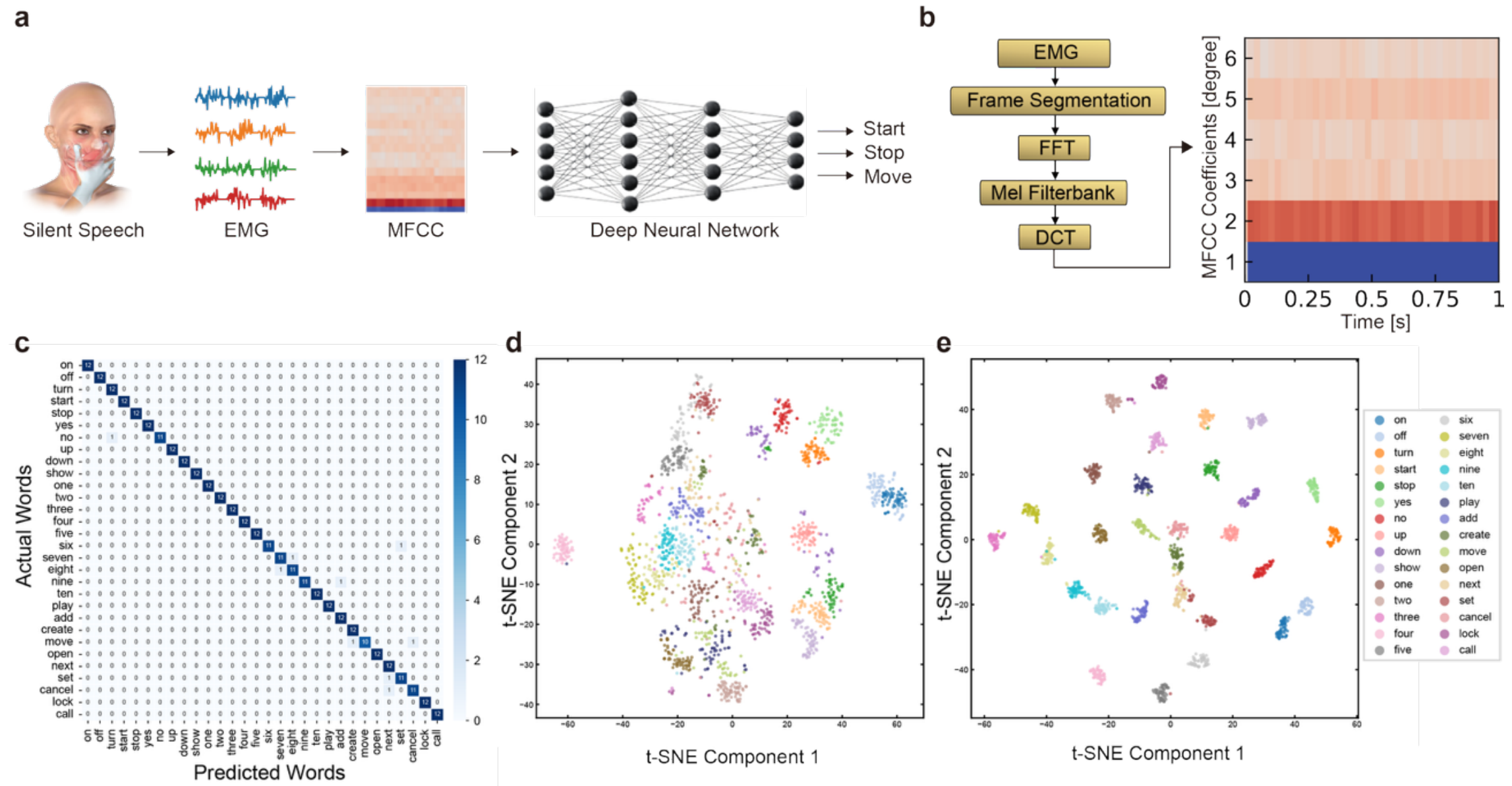

**Figure 4: Machine learning-based recognition of silent speech. a**, Acquisition of multichannel EMG signals and feature extraction using MFCCs. **b**, Preprocessing pipeline, including framing, FFT, filter bank analysis, and DCT for MFCC computation. **c**, Classification performance of the DNN, which recognized a 30-word vocabulary with an accuracy of 97.2 ± 1.3% across three subjects. **d**, Feature distribution before training, showing overlapping clusters of words. **e**, Feature distribution after DNN classification, demonstrating improved separability of silently articulated words.

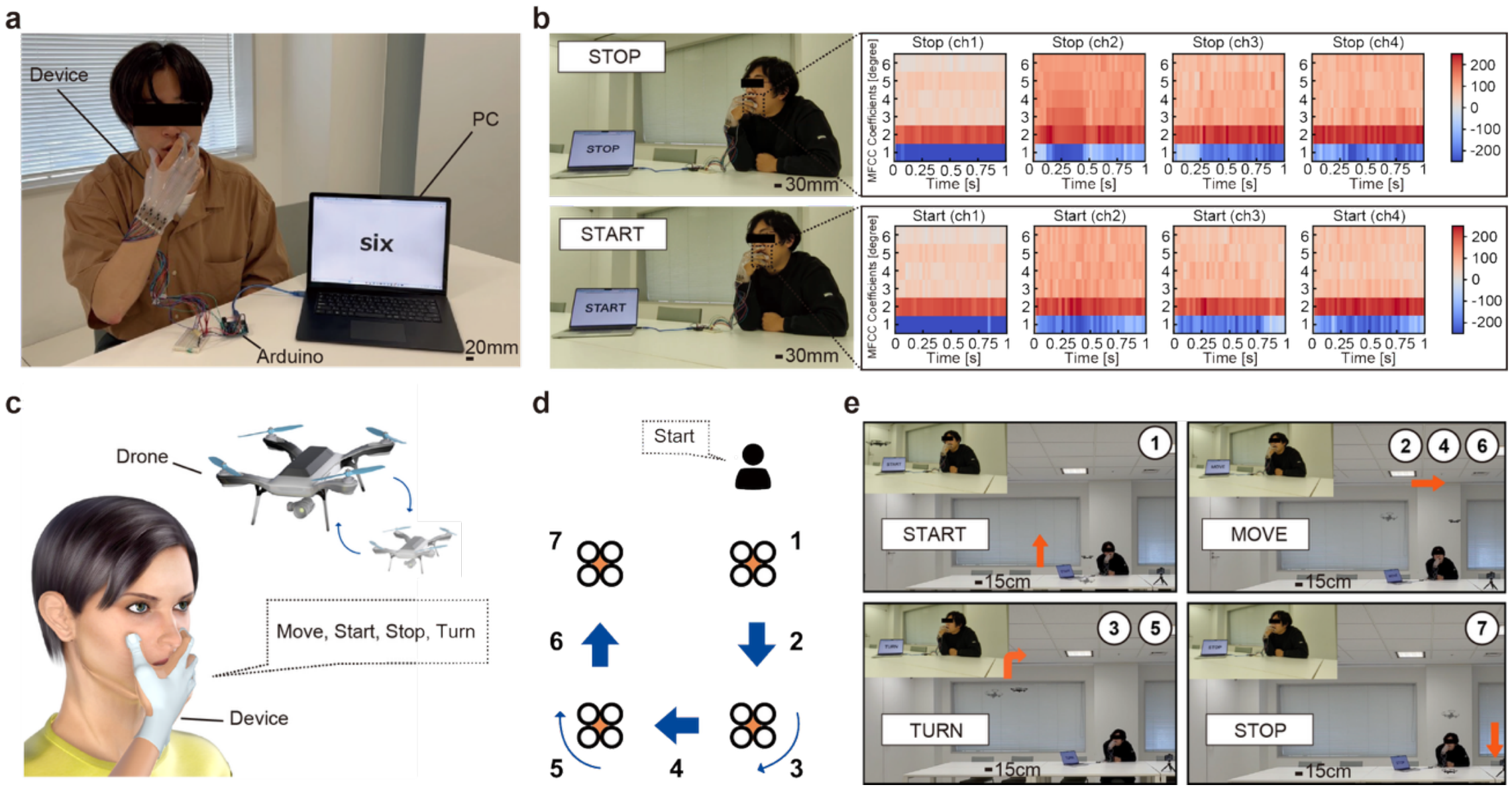

**Figure 5: Application to human–computer interaction through drone control. a**, Mechanism for silent speech recognition for drone control. Words classified by a pretrained model are displayed on a screen after the device is connected to an Arduino. **b**, Example MFCC features derived from four-channel EMG signals during articulation of "*Stop*" and "*Start*." **c**, Experimental setup for drone flight simulation using four commands ("*Start*," "*Move*," "*Turn*," and "*Stop*"). **d**, Flight scenario in which recognized commands are transmitted to the quadcopter via Wi-Fi, enabling real-time execution of operations (steps 1–7). **e**, Demonstration of drone takeoff, movement, turning, and landing executed solely through SSR using the device.

# Supplementary Information

## Soft Active EMG Interface for Machine Learning-Enabled Silent Speech Recognition

Yuta Kurotaki[1,2], Shusuke Yamakoshi[1], Reitaro Yoshida[2], Yutaka Isoda[1], Tamami Takano[1], Yuji Isano[1], Yusuke Miyake[2], Kentaro Kuribayashi[2], Hiroki Ota[1*]

[1]Department of Mechanical Engineering, Yokohama National University, 79-5, Tokiwadai, Hodogaya-Ku, Yokohama, Kanagawa 240-8501, Japan

[2]Pepabo Research and Development Institute, GMO Pepabo, Inc., 26-1, Sakuragaoka, Shibuya Ward, Tokyo 150-8512, Japan

*Corresponding author: Hiroki Ota, ota-hiroki-xm@ynu.ac.jp

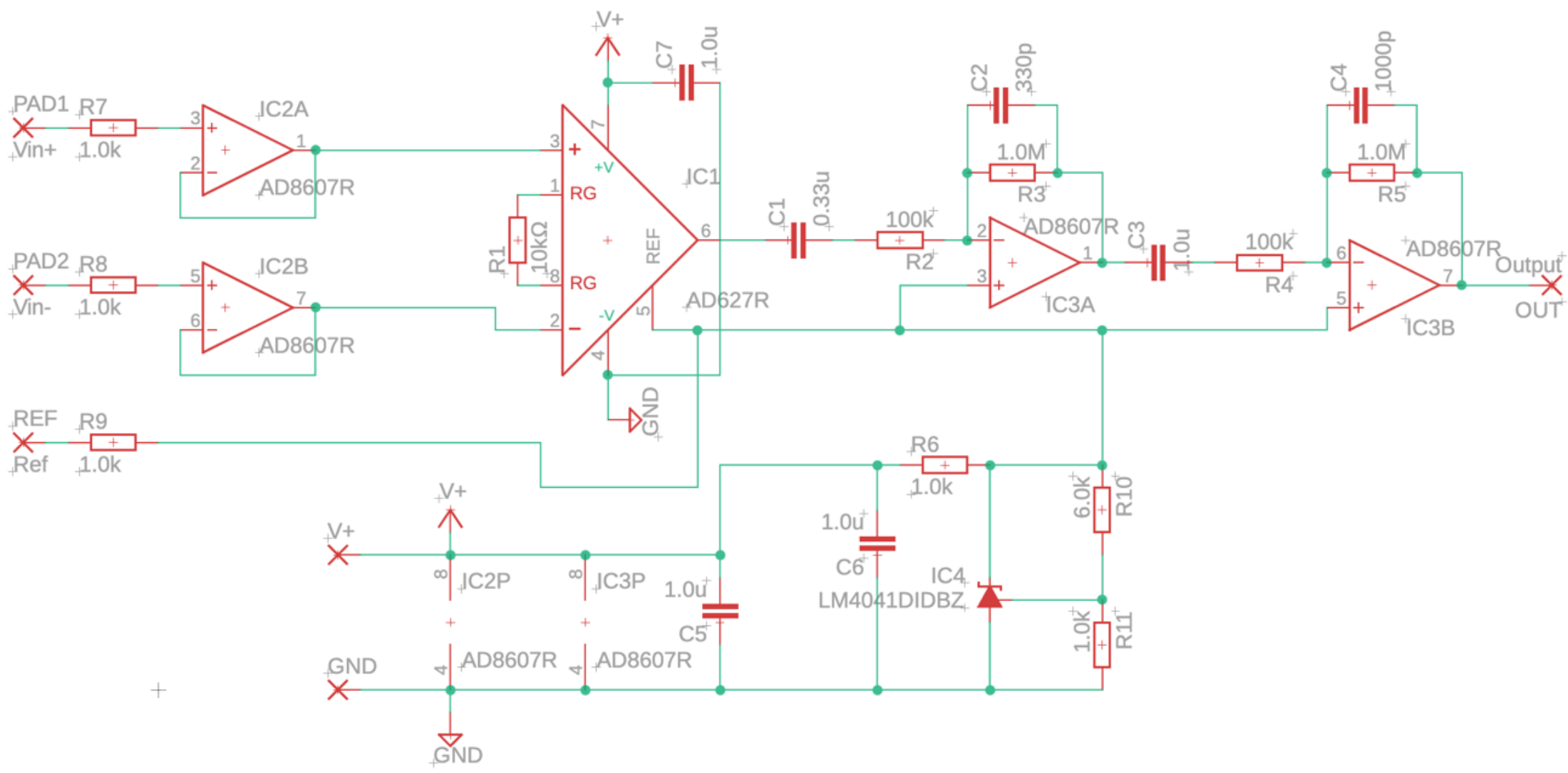

**Supplementary Figure 1: Circuit diagram of the EMG amplification system.** A voltage follower (AD8607, Analog Devices) was placed after the electrode, followed by a differential amplifier (AD627, Analog Devices) with a gain of 25. EMG signals were further amplified to 2,500 using subsequent operational amplifiers. Four identical circuits were integrated into the device.

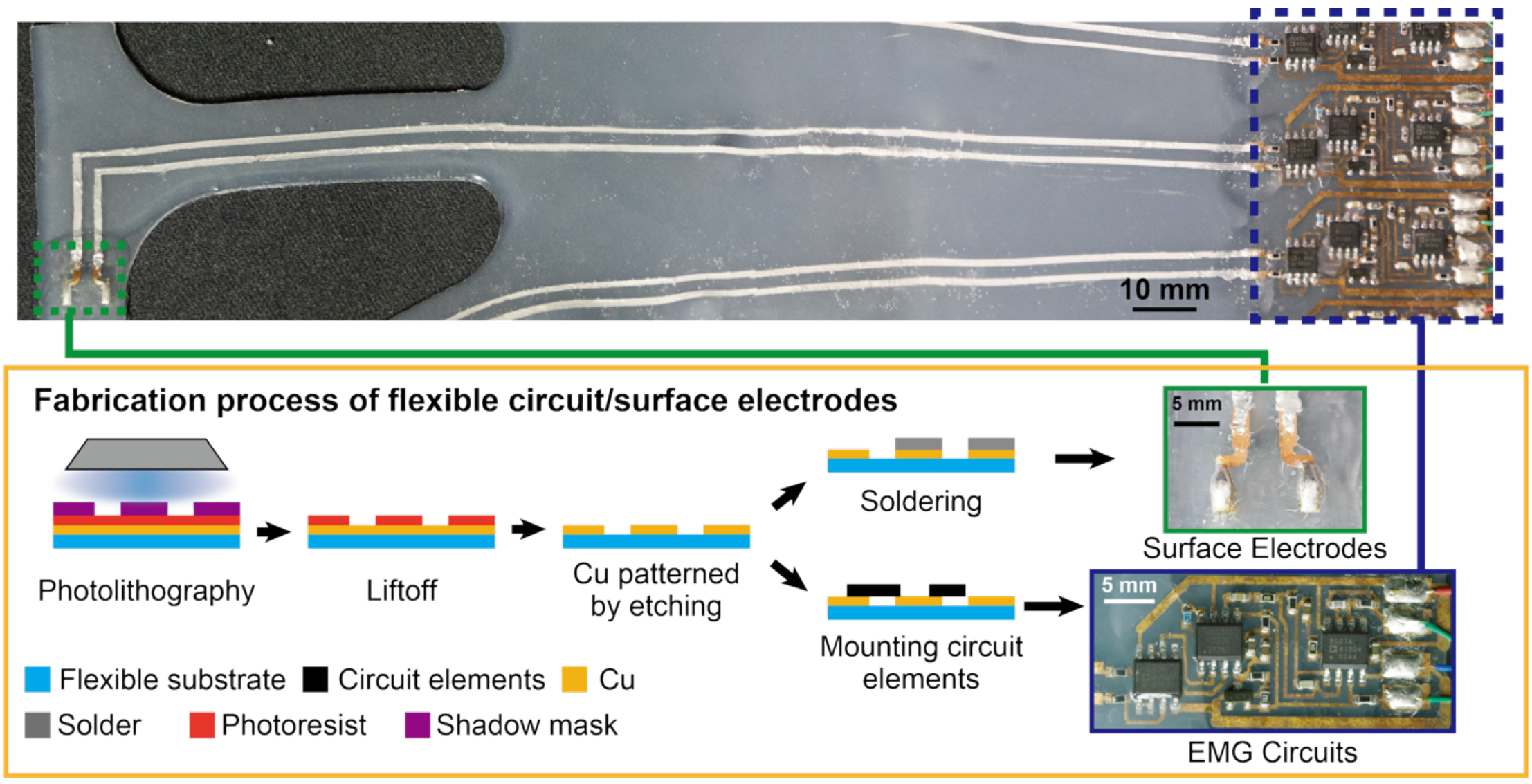

**Supplementary Figure 2: Fabrication process for the electrodes and circuit.** A transparent polyimide substrate coated on one side with copper foil (L71KTS 1012EDR T10, Arisawa Mfg. Co., Ltd.) was used. AZ photoresist was spin coated and dried on a hot plate at 100 °C to form a mask. The circuit pattern was then exposed and developed by etching to define the traces. Solder was applied to the electrode pads, and the EMG amplification circuit was populated with the required components.

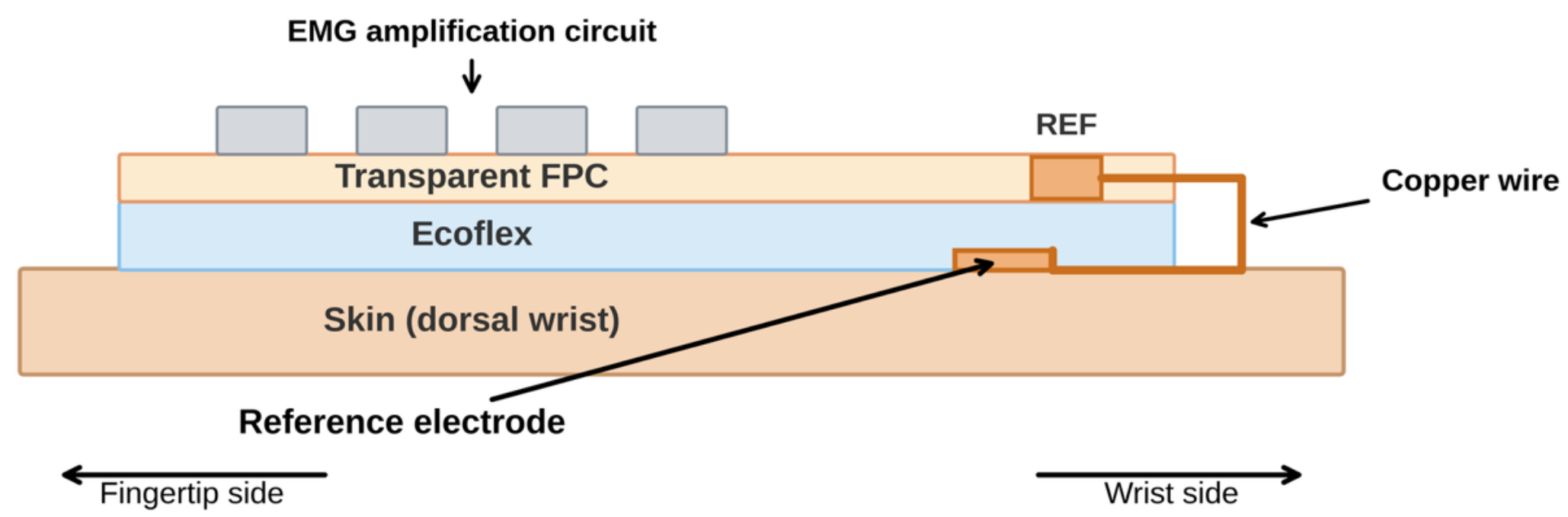

**Supplementary Figure 3: Cross-sectional schematic of the device showing the reference electrode location.** The device comprises a transparent FPC substrate carrying the EMG amplification circuit, encapsulated in Ecoflex. The reference electrode is routed from the REF terminal on the FPC via a copper wire through the Ecoflex layer to contact the skin on the dorsal side of the wrist. This placement takes advantage of the anatomical characteristics of the dorsal wrist, where tendons and bone are superficial rather than muscle bellies, resulting in minimal EMG contamination from local muscle activity.

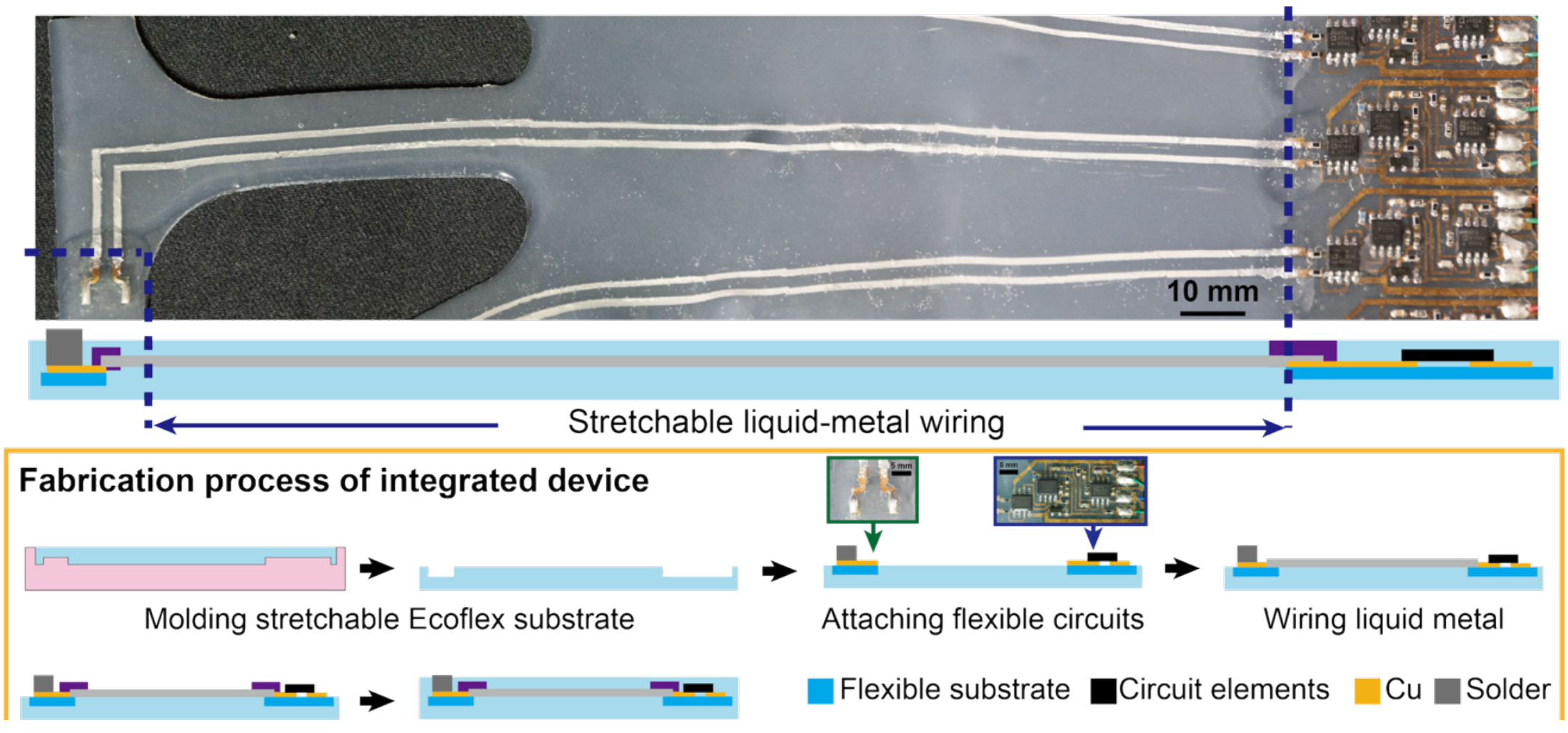

**Supplementary Figure 4: Fabrication process for the liquid metal wiring.** A mold produced via 3D printing was filled with silicone rubber Ecoflex 00-20 (Smooth-On; A:B = 1:1 by weight) and cured in an oven. The EMG amplification circuit board was fixed using Sil-Poxy adhesive (Smooth-On), and fingertip electrodes were attached in the same manner. A polyimide film was cut using a laser marker (Keyence) to form a stencil mask for liquid metal (LM) wiring. LM paste was applied through the stencil, and after mask removal, wiring connections between the fingertip electrodes and the EMG circuit were formed. The connection areas between the circuit board and the LM wiring were sealed with silicone rubber (KE-1606, Shin-Etsu Chemical) and cured. Finally, Ecoflex was coated over the entire structure and cured in an oven to

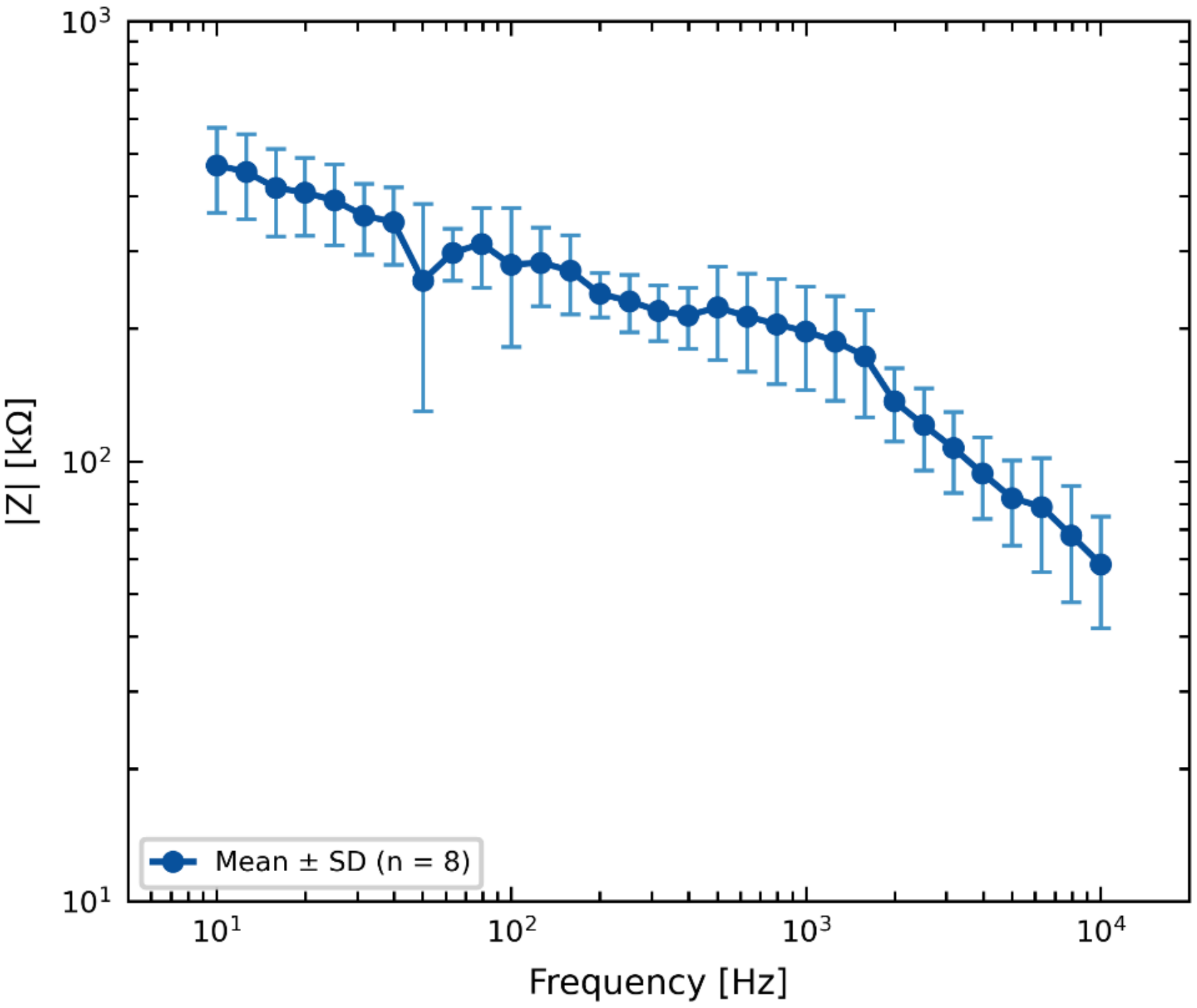

**Supplementary Figure 5: Electrode–skin impedance spectrum of the dry fingertip electrode.** Potentiostatic electrochemical impedance spectroscopy (PEIS) was performed using a BioLogic SP-300 potentiostat over a frequency range of 10 Hz to 10 kHz (Va = 10 mV, 10 points per decade). The plot shows the mean impedance magnitude |Z|, averaged over n = 8 independent measurements with the electrode in contact with forearm skin.

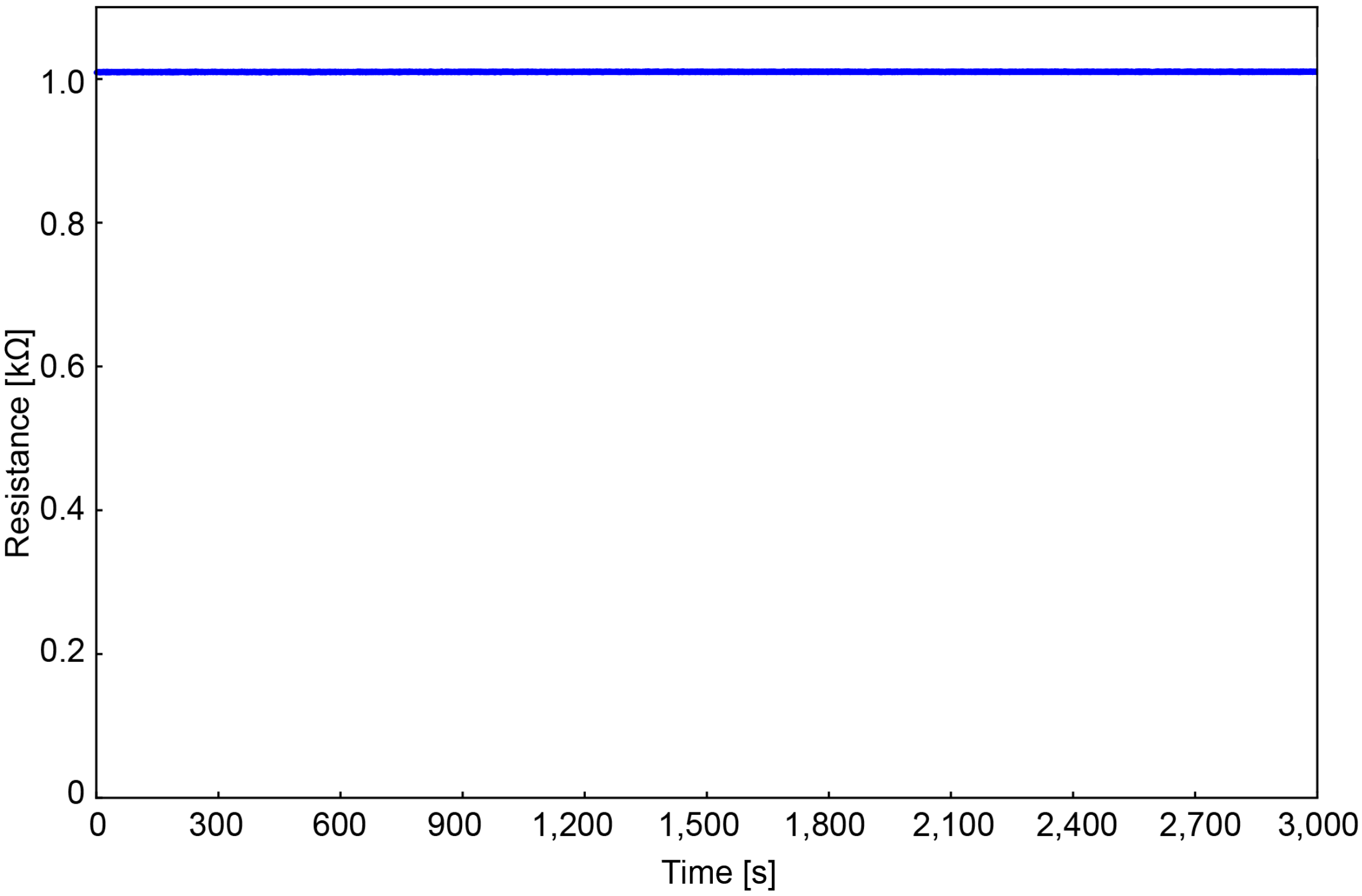

**Supplementary Figure 6: Resistance change during tensile testing of the electrodes.** Tensile tests were performed for 1,000 cycles at a strain of 100% using a tensile testing machine and an LCR meter, with resistance recorded at 1 s intervals. The resistance remained stable throughout the cycles, confirming the durability of the liquid metal wiring under repeated large deformation.

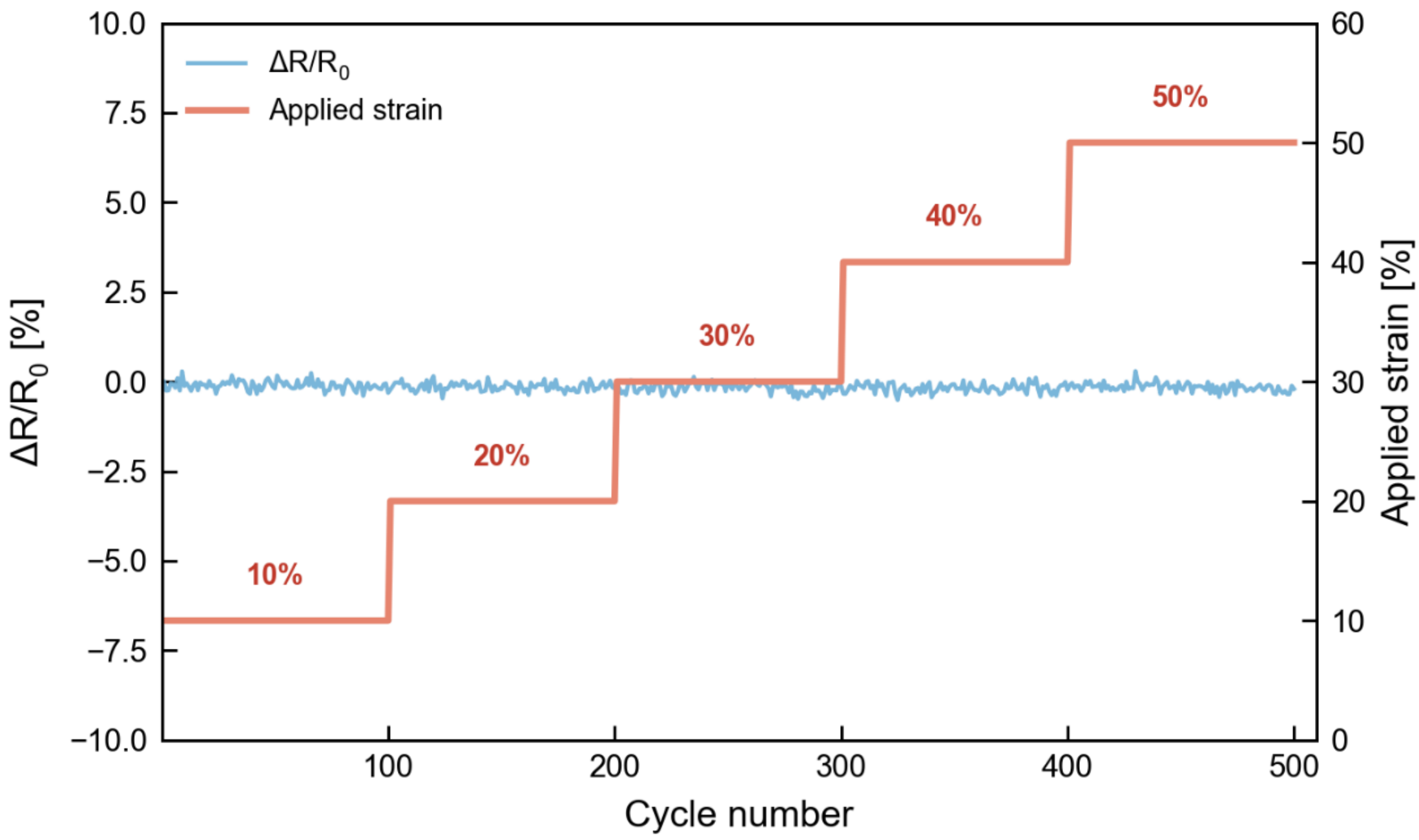

**Supplementary Figure 7: Resistance change of the liquid metal interconnects under various tensile strains.** The relative resistance change ($\Delta R/R_0$) of the electrode–circuit interconnection via liquid metal wiring was measured under stepwise tensile strains of 10%, 20%, 30%, 40%, and 50% (100 cycles each) using a tensile testing machine and an LCR meter, with resistance recorded at 1 s intervals. The resistance remained stable ($\Delta R/R_0 < \pm 0.5\%$) across all strain levels, confirming the mechanical durability of the liquid metal wiring under repeated deformation.

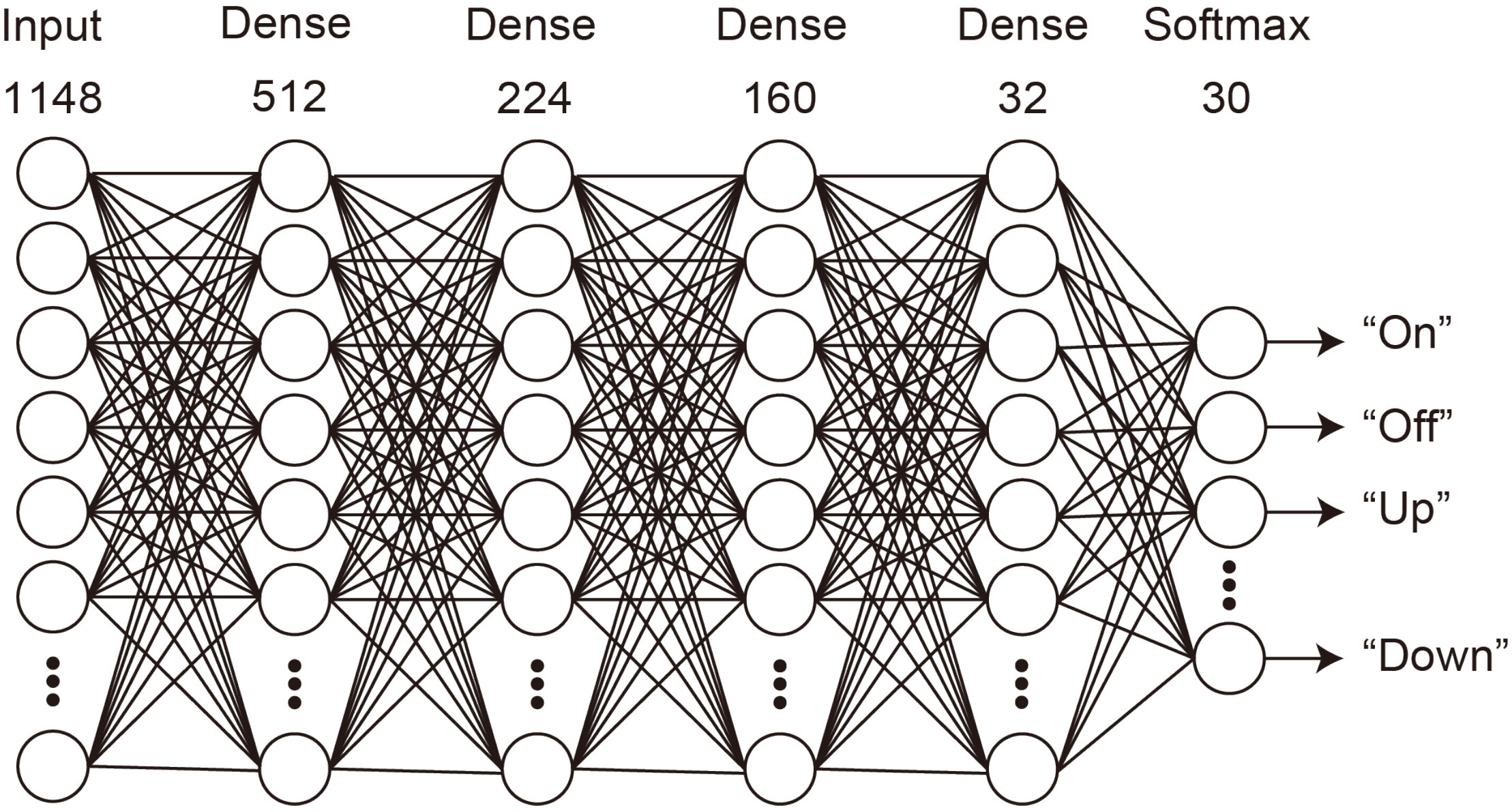

**Supplementary Figure 8: DNN model used in this study.** A fully connected DNN comprising multiple dense layers was employed for 30-word silent speech classification, with a softmax output layer for multiclass prediction. The architecture was determined through systematic hyperparameter optimization using Keras Tuner (Hyperband algorithm), in which the number of hidden layers, units per layer, dropout rate, and learning rate were explored across a broad search space.

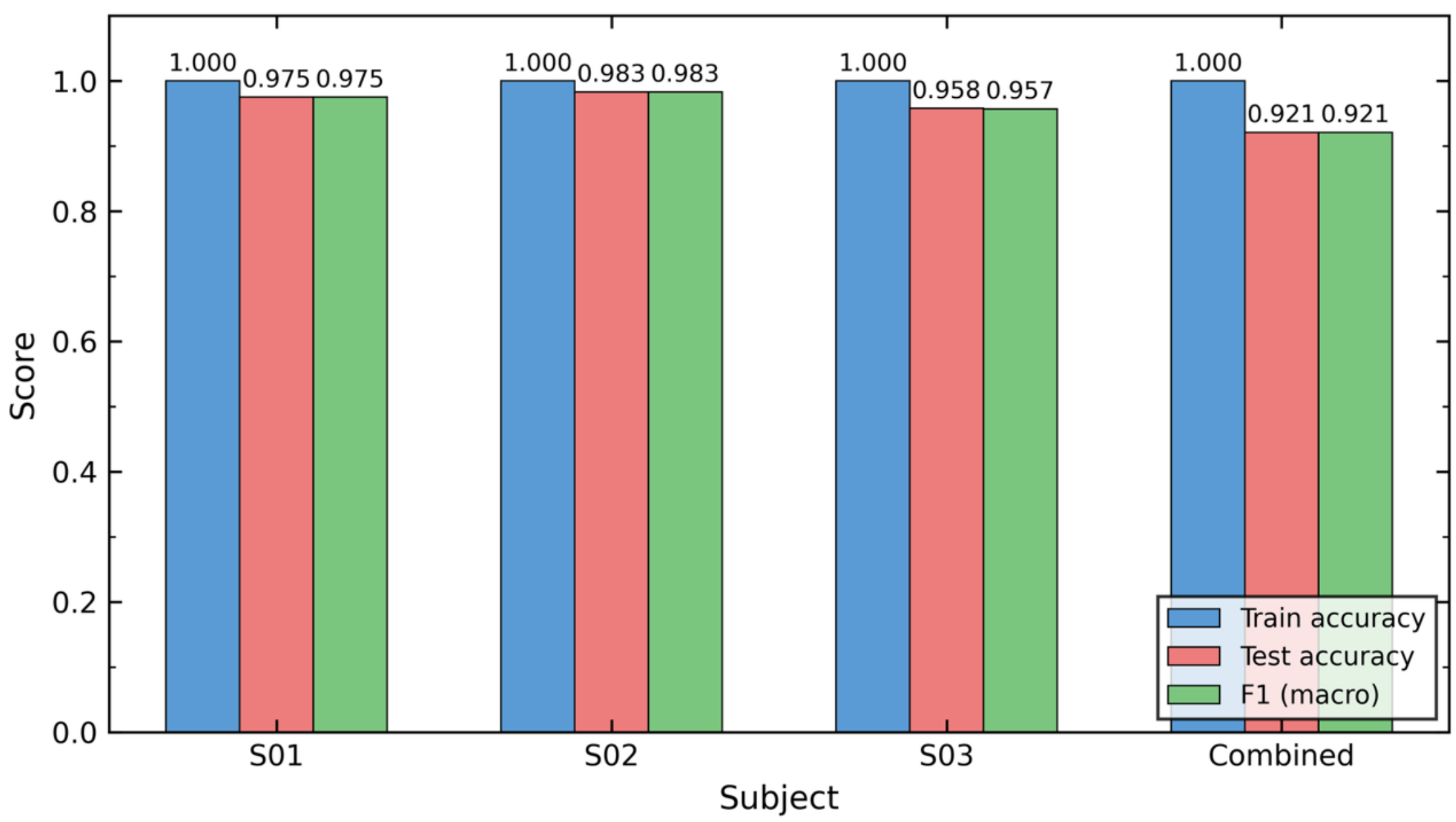

**Supplementary Figure 9: Comparison of training accuracy, test accuracy, and macro-averaged F1 score across models.** Training accuracy reached 1.000 for all conditions, as expected for a DNN with sufficient capacity trained on structured EMG patterns. Individual subject-dependent models (S01, S02, S03) achieved test accuracies of 0.958–0.983 (mean 0.972 ± 0.013), with train–test gaps of 1.7–4.2 percentage points. The close agreement between test accuracy and macro F1 across all conditions indicates balanced classification performance across all 30 word classes, with no evidence of reliance on majority-class predictions. The combined model, trained on pooled data from all three subjects (N = 5,400), achieved a test accuracy of 0.921 and a macro F1 score of 0.921, demonstrating cross-subject generalization despite increased inter-subject variability in EMG patterns.

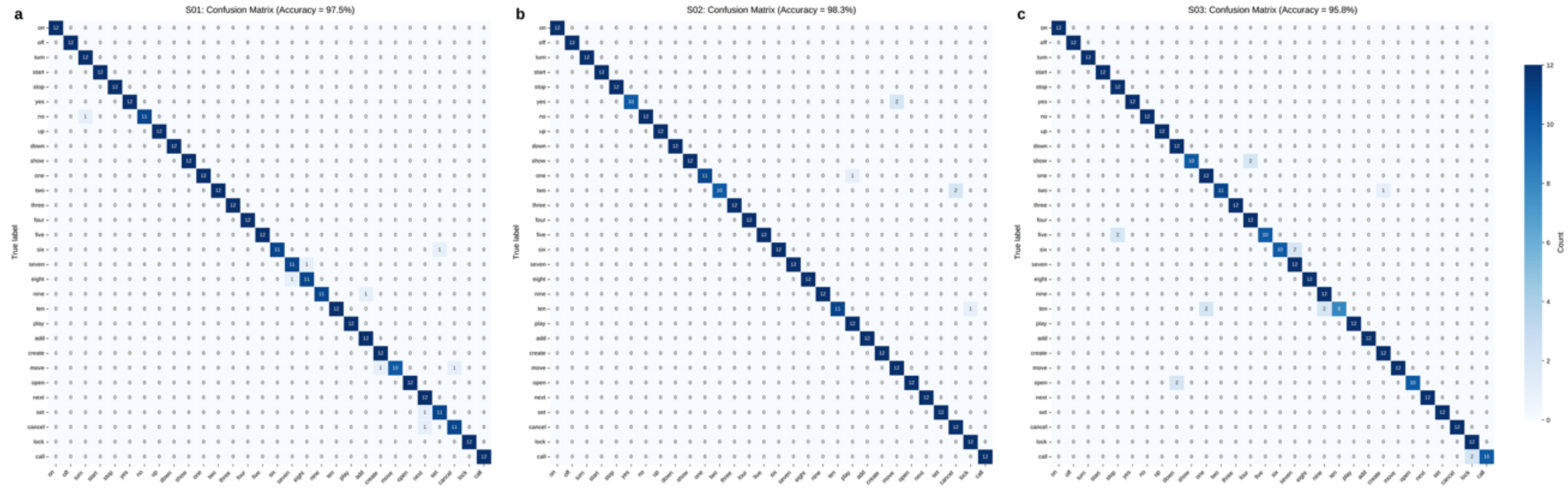

**Supplementary Figure 10: Confusion matrices of DNN-based silent speech recognition for individual participants.** a, Confusion matrix for participant S01. Rows indicate true word labels and columns indicate predicted labels; color intensity reflects the number of trials per cell, ranging from 0 (white) to 12 (dark blue). b, Confusion matrix for participant S02. The classification pattern is largely consistent with that of S01, confirming inter-subject reproducibility of the MFCC-based DNN model. c, Confusion matrix for participant S03. Diagonal dominance across all 30 words demonstrates that the model successfully discriminates the target vocabulary using perioral EMG signals acquired via fingertip dry electrodes.

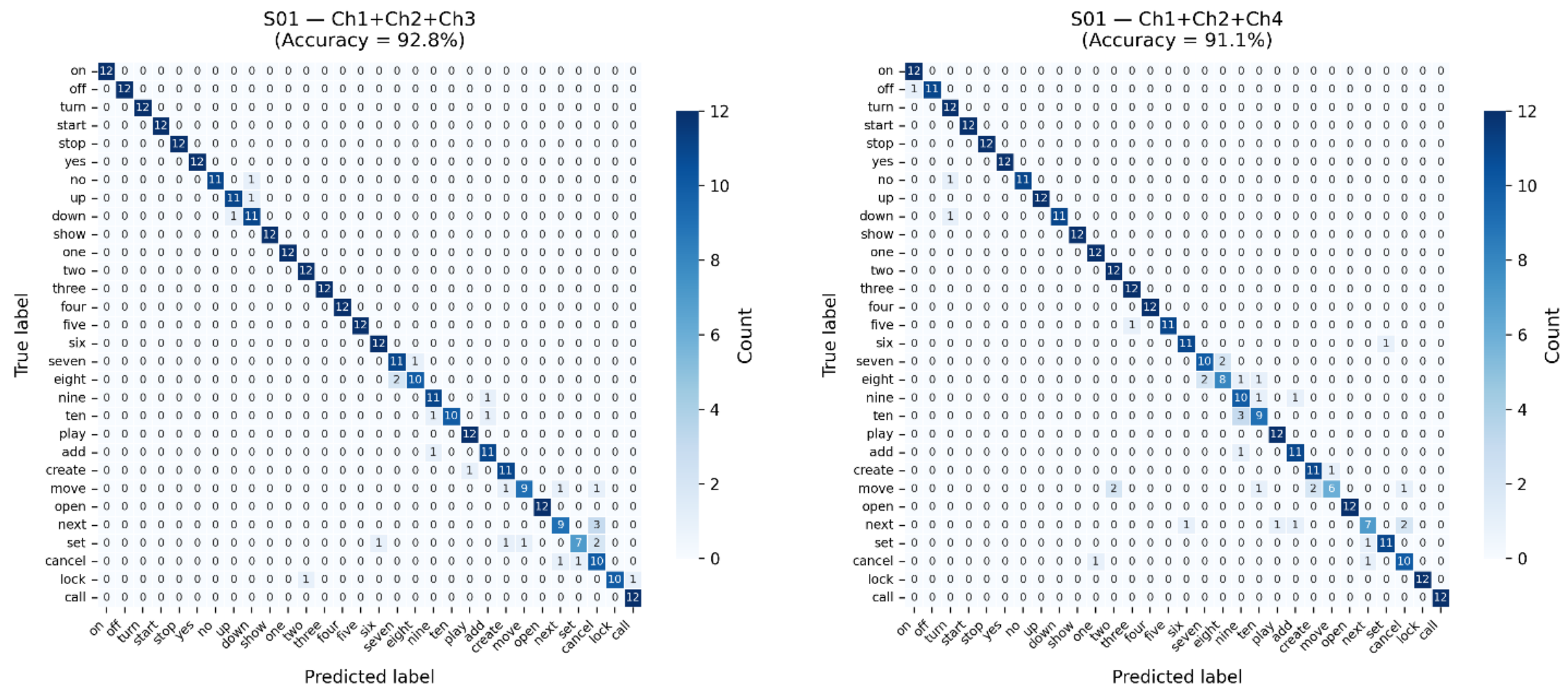

**Supplementary Figure 11: Comparison of confusion matrices for three-channel configurations.** The combination of Ch1, Ch2, and Ch3 yielded an accuracy of 92.8%, whereas the combination of Ch1, Ch2, and Ch4 yielded an accuracy of 91.1%.

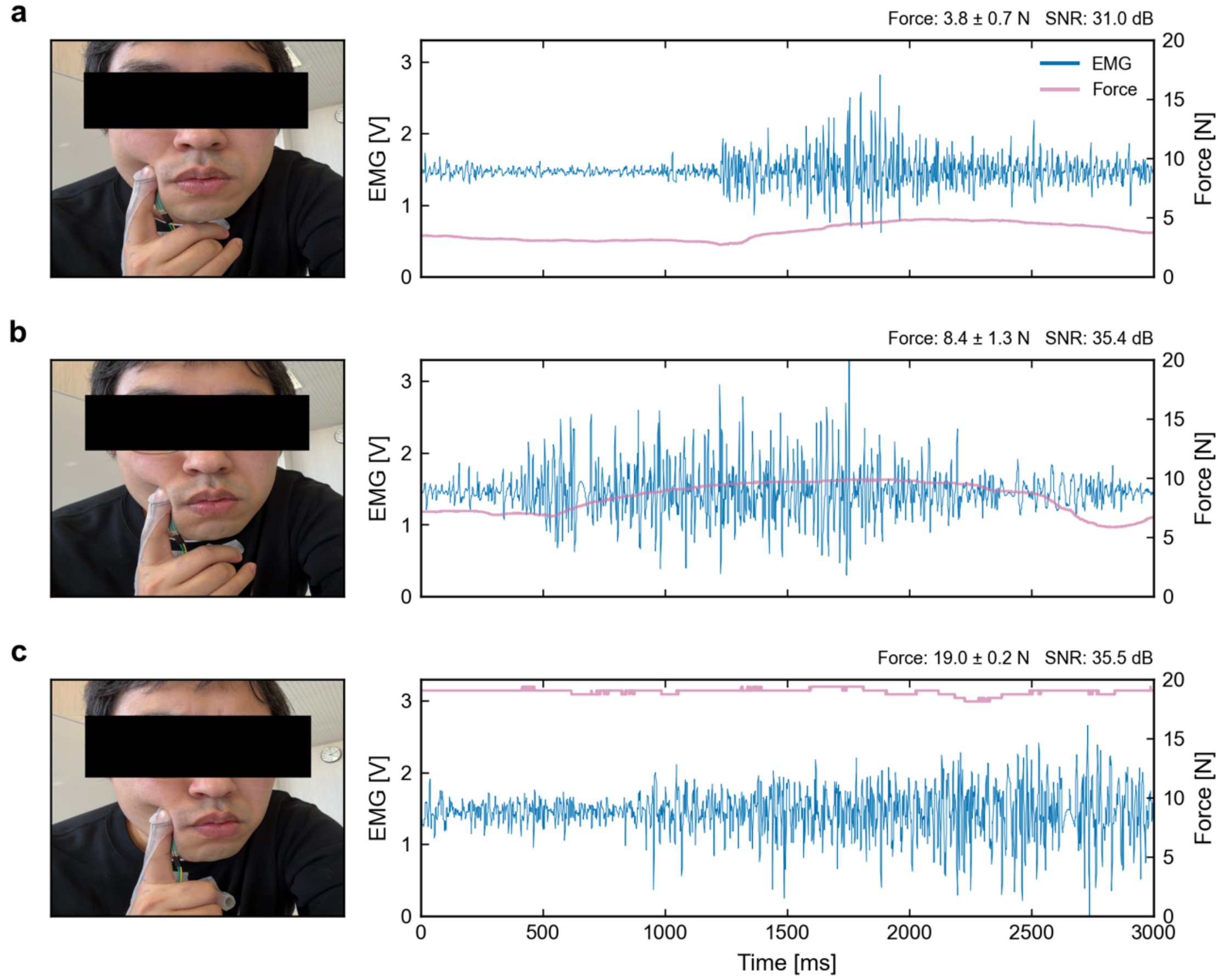

**Supplementary Figure 12: Surface EMG signals from the buccinator muscle during silent speech under three levels of applied pressure.** Each row shows a representative 3 s segment from the same recording session (subject S01, electrode fixed on the right buccinator muscle). Left panels show photographs of sensor placement during measurement. Right panels show EMG waveforms and simultaneous force sensor output (pink, right axis). a, Low-pressure condition (3.8 ± 0.7 N), SNR = 31.0 dB. b, Medium-pressure condition (8.4 ± 1.3 N), SNR = 35.4 dB. c, High-pressure condition (19.0 ± 0.2 N), SNR = 35.5 dB.

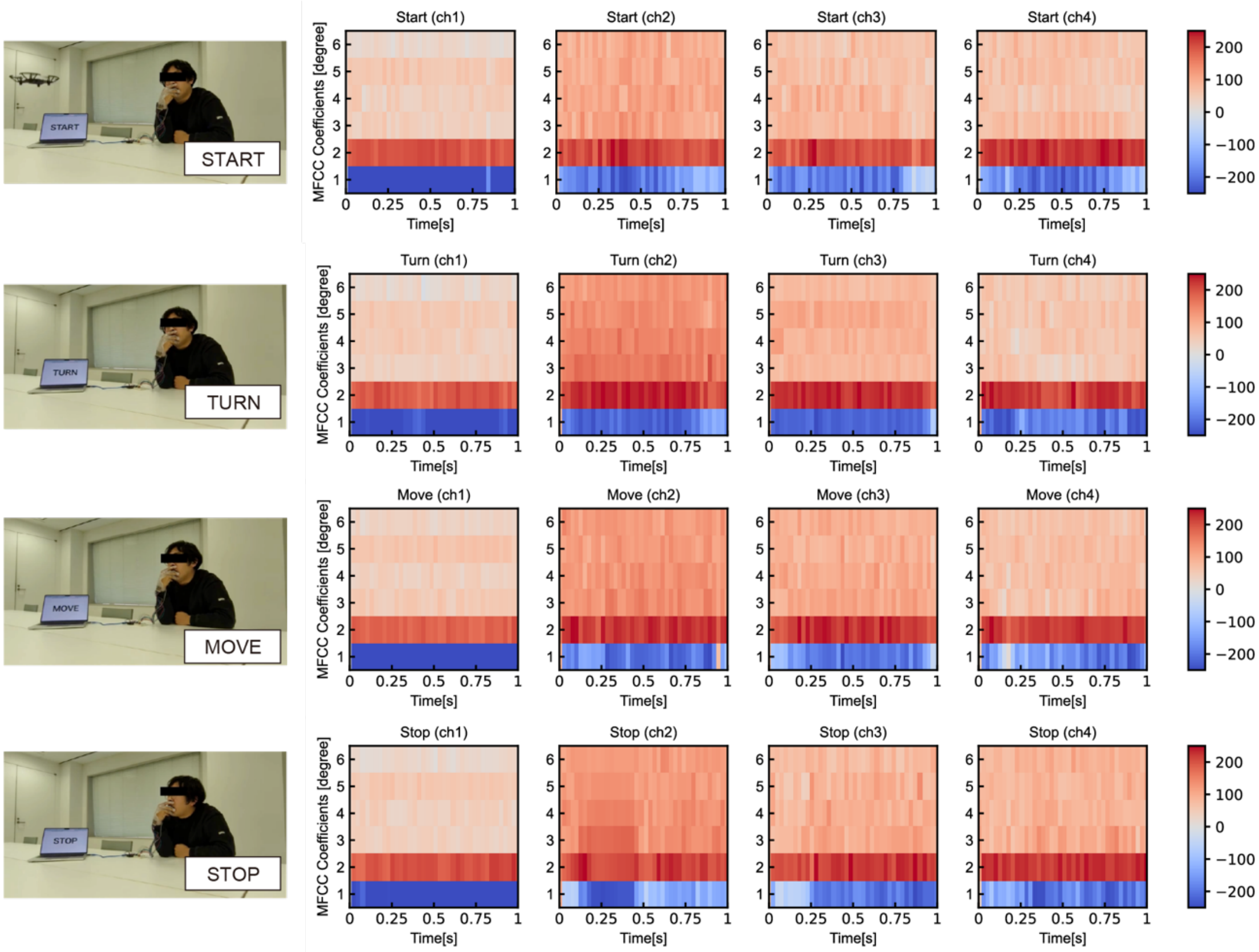

**Supplementary Figure 13: MFCCs for each word articulation.** MFCCs for each channel were calculated over 1-s intervals for four words ("START," "TURN," "MOVE," and "STOP").

Accuracy 91.1%

## Conformer Model

Input Layer
1148 dimensions

Dense (64)
73,536 params

Conformer Block 1
Self-Attention + Conv + FFN
263,872 params

Conformer Block 2
Self-Attention + Conv + FFN
263,872 params

Global Average Pooling
Temporal aggregation

Dense (96)
6,240 params

Output (30)
2,910 params

Accuracy 97.2%

## DNN Model

Input Layer
1148 dimensions

Dense (512)
588,288 params

Dense (224)
114,912 params

BatchNormalization

Dense (160)
36,000 params

BatchNormalization

Dense (32)
5,152 params

BatchNormalization

Output (30)
990 params

**Supplementary Figure 14: Comparison of Conformer and DNN models.** The Conformer and DNN models achieved classification accuracies of 91.1% and 97.2%, respectively.

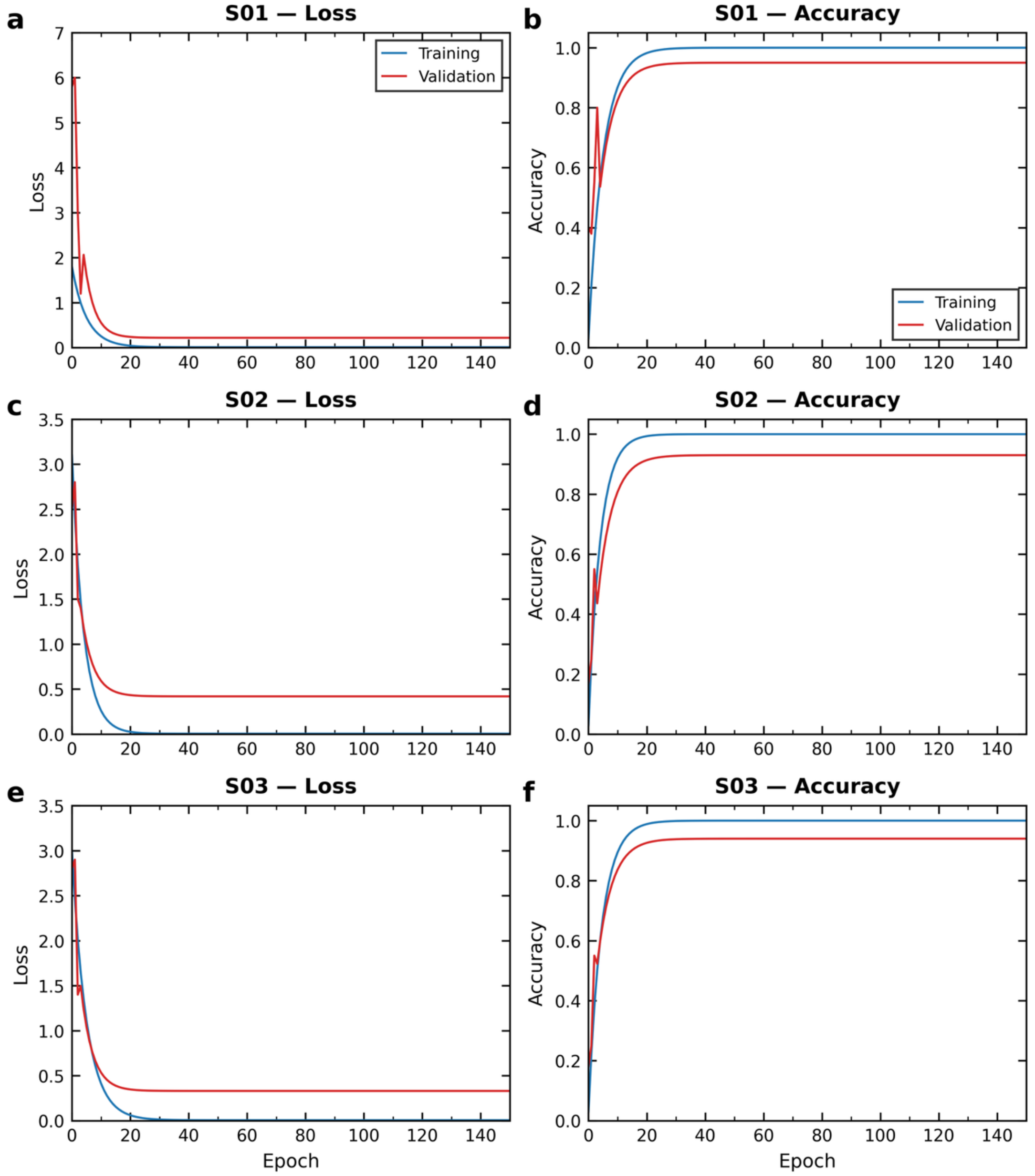

**Supplementary Figure 15: Training dynamics of the DNN classifier for individual subject-dependent models.** (a,b) Training and validation loss (a) and accuracy (b) over 150 epochs for subject S01 (N = 1,800). (c,d) Training and

**Supplementary Table 1: Comparison of accuracy by channel combination.**

| Channel combination | Accuracy |
|---|---|
| **1, 2, 3, 4** | **0.972** |
| 1, 2, 3 | 0.928 |
| 1, 2, 4 | 0.911 |
| 1, 4 | 0.900 |
| 1, 3 | 0.869 |
| 3, 4 | 0.867 |
| 1, 2 | 0.842 |
| 2, 3 | 0.814 |

[1]Hyperparameter optimization with K-fold cross-validation was performed only for the top three channel combinations (1,2,3,4; 1,2,3; and 1,2,4). Other combinations used default parameters.

**Supplementary Table 2: Comparison of silent speech recognition devices and their performance.**

| Study | Sensing method | Attachment method | Measurement timing | Visual obtrusiveness[a] | Ease of attachment[b] | Word recognition accuracy (%) | Ref. |
|---|---|---|---|---|---|---|---|
| *Kimura et al.* (2020) | Camera-based | Neck-mounted | Continuous | Slightly noticeable | Medium | 94.00 | [2] |
| *Vorontsova et al.* (2021) | EEG | Net cap | Continuous | Noticeable | Low | 84.51 | [10] |
| *Igarashi et al.* (2024) | Infrared sensor | Eyeglass frame | Continuous | Noticeable | Medium | 90.00[c] | [11] |
| *Kimura et al.* (2019) | Ultrasound | Handheld probe | Intent-driven | Noticeable | Low | 65.00[d] | [6] |
| *Srivastava et al.* (2022) | IMU | Earable device | Continuous | Slightly noticeable | Medium | 94.80 | [55] |
| *Srivastava et al.* (2023) | IMU | Earable device | Continuous | Unobtrusive | High | 97.07[e] | [50] |
| *Lu et al.* (2022) | TENGs | Mouth-mounted mask | Continuous | Noticeable | Medium | 94.50 | [15] |
| *Tang, C et al.* (2024) | Strain sensors | Neck-mounted | Continuous | Slightly noticeable | Medium | 95.25 | [44] |
| *Che et al.* (2024) | Magnetoelasticity | Tape fixation | Continuous | Slightly noticeable | Low | 94.68 | [54] |
| *Liu et al.* (2020) | sEMG | Tattoo-like | Continuous | Noticeable | Low | 89.60 | [52] |
| *Wang et al.* (2021) | sEMG | Tattoo-like | Continuous | Noticeable | Low | 92.64 | [53] |
| *Dong et al.* (2023) | Magnetic skin | Behind-the-ear patch | Continuous | Slightly noticeable | Medium | 93.20[g] | [47] |
| *Dong et al.* (2023) | sEMG | Facial patch (adhesive) | Continuous | Slightly noticeable | Low | 94.80[f] | [48] |
| *Dong et al.* (2025) | sEMG | Facial patch (adhesive) | Continuous | Slightly noticeable | Low | 91.50 | [49] |
| **This study (2025)** | **sEMG** | **Handheld (mouth)** | **Intent-driven** | **Unobtrusive** | **High** | **97.20** | **—** |

[a]Visual obtrusiveness assessed for facial mounting configurations: Unobtrusive, Slightly noticeable, Noticeable.
[b]Ease of attachment/detachment evaluated for practical deployment: High, Medium, Low.
EEG, electroencephalography; IMU, inertial measurement unit; TENGs, triboelectric nanogenerators; sEMG, surface electromyography.
[c]F-measure for 5 speech commands under silent speech condition; not classification accuracy.
[d]Recognition success ratio averaged over two users (Network 1 + Network 2; Table 1 of the original paper).
[e]Balanced accuracy (BAC) for user authentication task (Jawthenticate); not word recognition accuracy.
[f]LDA classifier, four-subject evaluation on 11 words from different viseme groups.
[g]Phoneme-level recognition accuracy (nine phonemes, average across subjects); not word-level recognition.

**Supplementary Table 3: Classification performance of the DNN model across subjects.** Summary of 5-fold cross-validation scores, held-out test accuracy, and macro-averaged F1 scores for each subject-dependent model (S01, S02, S03), the combined cross-subject model, and the mean across individual subjects. All models were trained for 150 epochs using a DNN architecture optimized with Keras Tuner.

| Experiment | N | 5-fold CV | Test accuracy | F1 (macro) | Individual fold scores |
|---|---|---|---|---|---|
| S01 | 1,800 | 0.934 | 0.975 | 0.975 | 0.944, 0.948, 0.924, 0.927, 0.926 |
| S02 | 1,800 | 0.899 | 0.983 | 0.983 | 0.899, 0.896, 0.938, 0.892, 0.872 |
| S03 | 1,800 | 0.914 | 0.958 | 0.957 | 0.906, 0.917, 0.924, 0.903, 0.921 |
| Combined | 5,400 | 0.836 | 0.921 | 0.921 | 0.853, 0.830, 0.854, 0.831, 0.846 |
| *Mean ± s.d. (S01–S03)* | — | *0.916 ± 0.018* | *0.972 ± 0.013* | *0.972 ± 0.013* | — |

N, number of samples; CV, cross-validation; s.d., standard deviation; F1 (macro), macro-averaged F1 score.
All models were trained for 150 epochs using a DNN architecture optimized via Keras Tuner.
S01–S03 denote individual subject-dependent models. Combined denotes a model trained on pooled data from all three subjects.
Training accuracy reached 1.000 for all conditions. Test accuracy and F1 (macro) were evaluated on a held-out test set.